\documentstyle[jair,twoside,11pt,theapa]{article}
\input epsf 

\jairheading{25}{2006}{529-576}{4/05}{4/06}
\ShortHeadings{Asynchronous Partial Overlay:  A New Algorithm for DCSP}
{Mailler \& Lesser}
\firstpageno{529}

\newcommand{\progstyle}[1]{
\begin{centering}
\prog{xx\=xx\=xx\=xx\=xx\=xx\= \kill #1}
\end{centering}
}

\newcommand{\prog}[1]{
\begin{centering} 
\fbox{\ \ \parbox{0pt}{
\begin{tabbing}#1 \end{tabbing}}\ \ \ }
\end{centering}  
}

\newcommand{\nnl}{\\[-0.02cm]}

\begin{document}

\newtheorem{defn}{Definition}
\newtheorem{property}{Property}

\title{Asynchronous Partial Overlay:  A New Algorithm for Solving Distributed Constraint Satisfaction Problems}

\author{\name Roger Mailler \email mailler@ai.sri.com \\
       \addr SRI International\\
       333 Ravenswood Dr\\
       Menlo Park, CA 94025 USA
       \AND
       \name Victor R. Lesser \email lesser@cs.umass.edu \\
       \addr University of Massachusetts, Department of Computer Science\\
       140 Governors Drive\\
       Amherst, MA 01003 USA}


\maketitle

\begin{abstract}
Distributed Constraint Satisfaction (DCSP) has long been considered an
important problem in multi-agent systems research.  This is because
many real-world problems can be represented as constraint satisfaction
and these problems often present themselves in a distributed
form.  In this article, we present a new complete, distributed
algorithm called {\em asynchronous partial overlay (APO)} for solving
DCSPs that is based on a cooperative mediation process.  The primary
ideas behind this algorithm are that agents, when acting as a
mediator, centralize small, relevant portions of the DCSP, that these
centralized subproblems overlap, and that agents increase the size of
their subproblems along critical paths within the DCSP as the problem
solving unfolds.  We present empirical evidence that shows that APO
outperforms other known, complete DCSP techniques.
\end{abstract}

\section{Introduction}
\label{Introduction}
The Distributed constraint satisfaction problem has become a very
useful representation that is used to describe a number of problems in
multi-agent systems including distributed resource allocation
\cite{conry91} and distributed scheduling \cite{sycara91}.  Some
researchers in cooperative multi-agent systems have focused on
developing methods for solving these problems that are based on one
key assumption.  Particularly, the agents involved in the problem
solving process are autonomous.  This means that the agents are only
willing to exchange information that is directly relevant to the
shared problem and retain the ability to refuse a solution when it
obviously conflicts with some internal goal.

These researchers believe that the focus on agent autonomy precludes
the use of centralization because it forces the agents to reveal all
of their internal constraints and goals which may, for reasons of
privacy or pure computational complexity, be impossible to achieve.
Several algorithms have been developed with the explicit purpose of
allowing the agents to retain their autonomy even when they are
involved in a shared problem which exhibits
interdependencies. Probably the best known algorithms that fit this
description can be found in the work of Yokoo et al. in the form of
distributed breakout (DBA) \cite{yokoo96dbo}, asynchronous
backtracking (ABT) \cite{yokoo92abt}, and asynchronous weak-commitment
(AWC) \cite{yokoo00review}.
  
Unfortunately, a common drawback to each of these algorithms is that
in an effort to provide the agents which {\bf complete privacy}, these
algorithms prevent the agents from making informed decisions about the
global effects of changing their local allocation, schedule, value,
etc.  For example, in AWC, agents have to try a value and wait for
another agent to tell them that it will not work through a $nogood$
message.  Because of this, agents never learn true reason why another
agent or set of agents is unable to accept the value, they only learn
that their value in combination with other values doesn't work.

In addition, these techniques suffer because they have complete
distribution of the control.  In other words, each agent makes
decisions based on its incomplete and often inaccurate view of the
world.  The result is that this leads to unnecessary thrashing in the
problem solving because the agents are trying to adapt to the behavior
of the other agents, who in turn are trying to adapt to them.
Pathologically, this behavior can be counter-productive to convergence
of the protocol\cite{selman03dsn}.

This iterative trial and error approach to discovering the implicit
and implied constraints within the problem causes the agents to pass
an exponential number of messages and actually reveals a great deal of
information about the agents' constraints and domain values
\cite{yokoocp02}.  In fact, in order to be complete, agents using AWC have to
be willing to reveal all of their \emph{shared} constraints and domain
values.  The key thing to note about this statement is that AWC still
allows the agents to retain their autonomy even if they are forced to
reveal information about the variables and constraints that form the
global constraint network.

In this paper, we present a \emph{cooperative mediation} based DCSP
protocol, called Asynchronous Partial Overlay (APO).  Cooperative
mediation represents a new methodology that lies somewhere between
centralized and distributed problem solving because it uses
dynamically constructed, partial centralization.  This allows
cooperative mediation based algorithms, like APO, to utilize the speed
of current state-of-the-art centralized solvers while taking advantage
of opportunities for parallelism by dynamically identifying relevant
problem structure.

APO works by having agents asynchronously take the role of mediator.
When an agent acts as a mediator, it computes a solution to a portion
of the overall problem and recommends value changes to the agents
involved in the \emph{mediation session}.  If, as a result of its
recommendations, it causes conflicts for agents outside of the
session, it links with them preventing itself from repeating the
mistake in future sessions.

Like AWC, APO provides the agents with a great deal of autonomy by
allowing anyone of them to take over as the mediator when they notice
an undesirable state in the current solution to the shared problem.
Further adding to their autonomy, agents can also ignore
recommendations for changing their local solution made by other
agents.  In a similar way to AWC, APO is both sound and complete when
the agents are willing to reveal the domains and constraints of their
shared variables and allows the agents to obscure the states,
domains, and constraints of their strictly local variables.

In the rest of this article, we present a formalization of the DCSP
problem (section \ref{APO:dcsp}).  In section \ref{assumptions}, we
describe the underlying assumptions and motivation for this work.  We
then present the APO algorithm (section \ref{APO:protocol}) and give
an example of the protocol's execution on a simple 3-coloring problem
(section \ref{APO:example}).  We go on to give the proofs for the
soundness and completeness of the algorithm (section \ref{APO:proof}).
In section \ref{APO:evaluation}, we present the results of extensive
testing that compares APO with AWC within the distributed graph
coloring domain and the complete compatibility version of the
SensorDCSP domain
\cite{bejar01distributed} across a variety of metrics including number
of cycles, messages, bytes transmitted, and serial runtime.  In each
of these cases, we will show that APO significantly outperforms AWC
\cite{yokoo95awc,hirayama00learning}. Section \ref{summary} summarizes
the article and discusses some future research directions.

\section{Distributed Constraint Satisfaction}
\label{APO:dcsp}
A Constraint Satisfaction Problem (CSP) consists of the following:
\begin{itemize}
\item a set of {\it n} variables $V = \{x_1,\ldots,x_n\}$.
\item discrete, finite domains for each of the variables $D = \{D_1,\ldots,D_n\}$.
\item a set of constraints $R = \{R_1,\ldots,R_m\}$ where each $R_i(d_{i1},\ldots,d_{ij})$ is a predicate on the Cartesian product $D_{i1}\times \cdots \times D_{ij}$ that returns true iff the value assignments of the variables satisfies the constraint.
\end{itemize}

The problem is to find an assignment $A=\{d_1,\ldots,d_n|d_i \in
D_i\}$ such that each of the constraints in $R$ is satisfied.  CSP has
been shown to be NP-complete, making some form of search a necessity.

In the distributed case, DCSP, using variable-based decomposition,
each agent is assigned one or more variables along with the
constraints on their variables.  The goal of each agent, from a local
perspective, is to ensure that each of the constraints on its
variables is satisfied.  Clearly, each agent's goal is not independent
of the goals of the other agents in the system.  In fact, in all but
the simplest cases, the goals of the agents are strongly interrelated.
For example, in order for one agent to satisfy its local constraints,
another agent, potentially not directly related through a constraint,
may have to change the value of its variable.

In this article, for the sake of clarity, we restrict ourselves to the
case where each agent is assigned a single variable and is given
knowledge of the constraints on that variable.  Since each agent is
assigned a single variable, we will refer to the agent by the name of
the variable it manages.  Also, we restrict ourselves to considering
only binary constraints which are of the form
$R_i(d_{i1},d_{i2})$. Since APO uses centralization as its core, it is
easy to see that the algorithm would work if both of these
restrictions are removed.  This point will be discussed as part of the
algorithm description in section \ref{APO:remove}.
 
Throughout this article, we use the term \emph{constraint graph} to
refer to the graph formed by representing the variables as nodes and
the constraints as edges.  Also, a variable's \emph{neighbors} are the
variables with which it shares constraints.

\pagebreak

\section{Assumptions and Motivation}
\label{assumptions}

\subsection{Assumptions}
The following are the assumptions made about the environments and
agents for which this protocol was designed:

\begin{enumerate}

\item Agents are situated, autonomous, computing entities.  As such, they are capable of sensing their environment, making local decisions based on some model of intentionality, and acting out their decisions.  Agents are rationally and resource bounded.  As a result, agents must communicate to gain information about each others state, intentions, decisions, etc.

\item Agents within the multi-agent system share one or more joint goals.  In this paper, this goal is Boolean in nature stemming from the DCSP formulation.

\item Because this work focuses on cooperative problem solving, the agents are cooperative.  This does not necessarily imply they will share all of their state, intentions, etc. with other agents, but are, to some degree, willing to exchange information to solve joint goals.  It also does not imply that they will change their intentions, state, or decisions based on the demands of another agent.  Agents still maintain their autonomy and the ability to refuse or revise the decisions of other agents based on their local state, intentions, decisions, etc.

\item Each agent has the capability of computing solutions to the joint goal based on their potentially limited rationality.  This follows naturally from the ability of agents to make their own decisions, i.e., every agent is capable of computing a solution to its own portion of the joint goal based on its own desires.

\end{enumerate}

\subsection{Motivation for Mediation-Based Problem Solving}
\label{motivation}
The Websters dictionary defines the act of mediating as follows: 
 
\begin{quote} 
{\bf Mediate}: 1.  to act as an intermediary; especially to work with
opposing sides in order to resolve (as a dispute) or bring about (as a
settlement).\ 2.\ \ to bring about, influence, or transmit by acting
as an intermediate or controlling agent or
mechanism. \cite{dictionary}
\end{quote}

By its very definition, mediation implies some degree of centralizing
a shared problem in order for a group of individuals to derive a
conflict free solution.  Clearly in situations where the participants
are willing (cooperative), mediation is a powerful paradigm for
solving disputes.  It's rather strange, considering this, that very
little has been done on looking at mediation as a cooperative method
for solving DCSPs.

Probably, the earliest mediation-based approach for solving
conflicts amongst agents in an airspace management application\cite{cammarata83}.  This
work investigates using various conflict resolution strategies to
deconflict airspace in a distributed air traffic control system.  The
author proposes a method for solving disputes where the involved
agents elect a leader to solve the problem.  Once elected, the leader
becomes responsible for recognizing the dispute, devising a plan to
correct it, and acting out the plan.  Various election schemes are
tested, but unfortunately, the leader only has authority to modify its
own actions in order to resolve the conflicts.  This obviously leads
to situations where the plan is suboptimal.

In \cite{hayden99architectural}, the authors describe mediation as one
of a number possible coordination mechanisms.  In this work,
the mediator acts as a intermediary between agents and can also act to
coordinate their behavior.  As an intermediary, the mediator routes
messages, provides directory services, etc. This provides for a loose
coupling of the agents, since they only need to know the mediator.
The mediator can also act to coordinate the agents behavior if they
have tight interdependencies.  

Most of the research work on mediation-based problem solving involved
settling disputes between competitive or semi-competitive agents.
Probably one of the best examples of using mediation in this manner
can be found in the PERSUADER system\cite{sycara88}. PERSUADER was designed to settle conflicts
between adversarial parties that are involved in a labor dispute.
PERSUADER uses case-based reasoning to suggest concessions in order to
converge on a satisfactory solution.  Another example of using
mediation in this way can be found in in a system called Designer Fabricator Interpreter (DFI) \cite{werkman90}.  In
DFI, mediation is used to resolve conflicts as one of a series of
problem solving steps.  Whenever the first step fails, in this case
iterative negotiation, a mediator agent steps in and tries to convince
the agents to relax their constraints.  If this fails, the mediator
mandates a final solution.

There may be several reasons why mediation has not been more deeply
explored as a cooperative problem solving method.  First, researchers
have focused strongly on using distributed computing as a way of
exploiting concurrency to distribute the computation needed to solve
hard problems \cite{rao93}.  Because of this, even partially and/or
temporarily centralizing sections of the problem can be viewed as
contradictory to the central goal. Second, researchers have often
claimed that part of the power of the distributed methods lies in the
ability of the techniques to solve problems that are naturally
distributed.  For example, supply chain problems generally have no central
monitoring authority.  Again, directly sharing the reasons why a particular choice is made in the form of a constraint can seem to contradict the
use of distributed methods.  Lastly, researchers often claim that for
reasons of privacy or security the problem should be solved in a
distributed fashion.  Clearly, sharing information to solve a problem
compromises an agents ability to be private and/or violates its
security in some manner.
  
\begin{figure}
  \epsfysize=0.7in
  \hspace*{\fill}
  \epsffile{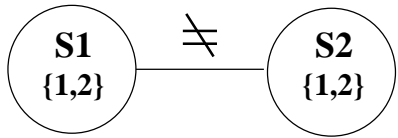}
  \hspace*{\fill}
  \caption{A simple distributed problem with two variables.}
  \label{simple-prob}
\end{figure}

Although, parallelism, natural distribution, security, and privacy,
may all seem like good justifications for entirely distributed problem
solving, in actuality, whenever a problem has interdependencies
between distributed problem solvers, some degree of centralization and
information sharing must take place in order to derive a conflict-free
solution.  

Consider, as a simple example, the problem in figure
\ref{simple-prob}.  In this figure, two problem solvers, each with one
variable, share the common goal of having a different value from one
another.  Each of the agents has only two allowable values:
\{1, 2\}.  Now, in order to solve this problem, each agent must
individually decide that they have a different value from the other
agent.  To do this, at the very least, one agent must transmit its
value to the other.  By doing this, it removes half of its privacy (by
revealing one of its possible values), eliminates security (because
the other agent could make him send other values by telling him that
this value is no good), and partially centralizes the problem solving
(agent S2 has to compute solutions based on the solution S1 presented
and decide if the problem is solved and agent S1 just relies on S2 to
solve it.)  In even this simple example, achieving totally distributed
problem solving is impossible.
 
In fact, if you look at the details of any of the current approaches
to solving DCSPs, you will observe a significant amount of
centralization occurring.  Most of these approaches perform this
centralization incrementally as the problem solving unfolds in an
attempt to restrict the amount of internal information being shared.
Unfortunately, on problems that have interdependencies among the
problem solvers, revealing an agent's information (such as the
potential values of their variables) is unavoidable.  In fact,
solutions that are derived when one or more agents conceals all of the
information regarding a shared constraint or variable are based on
incomplete information and therefore may not always be sound.
 
It follows then, that since you cannot avoid some amount of
centralization, mediation is a natural method for solving problems
that contain interdependencies among distributed problem solvers.

\section{Asynchronous Partial Overlay}
As a cooperative mediation based protocol, the key ideas behind the
creation of the APO algorithm are
\begin{itemize}
\item Using mediation, agents can solve subproblems of the DCSP using internal search.
\item These local subproblems can and should overlap to allow for more rapid convergence of the problem solving.
\item Agents should, over time, increase the size of the subproblem they work on along critical paths within the CSP.  This increases the overlap with other agents and ensures the completeness of the search.
\end{itemize} 
 
\subsection{The Algorithm}
\label{APO:protocol}
Figures \ref{apo_init}, \ref{apo_ok}, \ref{apo_mediator},
\ref{apo_choose}, and \ref{apo_receive} present the basic APO
algorithm.  The algorithm works by constructing a $good\_list$ and
maintaining a structure called the $agent\_view$.  The $agent\_view$
holds the names, values, domains, and constraints of variables to
which an agent is linked.  The $good\_list$ holds the names of the
variables that are known to be connected to the owner by a path in the
constraint graph.

As the problem solving unfolds, each agent tries to solve the
subproblem it has centralized within its $good\_list$ or determine
that it is unsolvable which indicates the entire global problem is
over-constrained.  To do this, agents take the role of the mediator
and attempt to change the values of the variables within the mediation
session to achieve a satisfied subsystem.  When this cannot be
achieved without causing a violation for agents outside of the
session, the mediator links with those agents assuming that they are
somehow related to the mediator's variable.  This process continues
until one of the agents finds an unsatisfiable subsystem, or all of
the conflicts have been removed.

In order to facilitate the problem solving process, each agent has a
dynamic priority that is based on the size of their $good\_list$ (if
two agents have the same sized $good\_list$ then the tie is broken
using the lexicographical ordering of their names).  Priorities are
used by the agents to decide who mediates a session when a conflicts
arises.  Priority ordering is important for two reasons.  First,
priorities ensure that the agent with the most knowledge gets to make
the decisions.  This improves the efficiency of the algorithm by
decreasing the effects of myopic decision making.  Second, priorities
improve the effectiveness of the mediation process because lower
priority agents expect higher priority agents to mediate.  This
improves the likelihood that lower priority agents will be available
when a mediation request is sent.

\begin{figure*}
\centerline{
\hbox{
\progstyle{
procedure {\bf initialize}\nnl
\> $d_i \leftarrow random\ d \in D_i$; \nnl
\> $p_i \leftarrow sizeof(neighbors) + 1$; \nnl
\> $m_i \leftarrow$ {\bf true}; \nnl
\> $mediate \leftarrow$ {\bf false}; \nnl
\> add $x_i$ to the $good\_list$; \nnl
\> send ({\bf init}, ($x_i, p_i, d_i, m_i, D_i, C_i$)) to neighbors; \nnl
\> $initList \leftarrow$ neighbors; \nnl
{\bf end initialize};\nnl \nnl
{\bf when received (init, ($x_j, p_j, d_j, m_j, D_j, C_j$))} {\bf do} \nnl
\> Add ($x_j, p_j, d_j, m_j, D_j, C_j$) to {\it agent\_view}; \nnl
\> {\bf if} $x_j$ is a neighbor of some $x_k \in good\_list$ {\bf do} \nnl
\>\> add $x_j$ to the {\it good\_list}; \nnl
\>\> add all $x_l \in agent\_view \land x_l \notin good\_list$  \nnl
\>\>\>that can now be connected to the {\it good\_list}; \nnl
\>\> $p_i \leftarrow sizeof(good\_list)$; \nnl
\> {\bf end if}; \nnl
\> {\bf if} $x_j \notin initList$ {\bf do} \nnl
\>\> send ({\bf init}, ($x_i, p_i, d_i, m_i, D_i, C_i$)) to $x_j$; \nnl
\> {\bf else} \nnl
\>\> remove $x_j$ from {\it initList}; \nnl
\>{\bf check\_agent\_view}; \nnl
{\bf end do};
}
}
}
\caption{The APO procedures for initialization and linking.}
\label{apo_init}
\end{figure*}

\subsubsection{Initialization (Figure \ref{apo_init})}
\label{APO:init}
On startup, the agents are provided with the value (they pick it
randomly if one isn't assigned) and the constraints on their variable.
Initialization proceeds by having each of the agents send out an
``init'' message to its neighbors.  This initialization message
includes the variable's name ($x_i$), priority ($p_i$), current
value($d_i$), the agent's desire to mediate ($m_i$), domain ($D_i$),
and constraints ($C_i$).  The array {\it initList} records the names
of the agents that initialization messages have been sent to, the
reason for which will become immediately apparent.

\begin{figure}
\centerline{
\hbox{
\progstyle{
{\bf when received} ({\bf ok?}, ($x_j, p_j, d_j, m_j$)) {\bf do} \nnl
\> update {\it agent\_view} with $(x_j, p_j, d_j, m_j)$; \nnl
\> {\bf check\_agent\_view}; \nnl
{\bf end do}; \nnl \nnl
procedure {\bf check\_agent\_view}\nnl
\> {\bf if} \ $initList \neq \emptyset$ {\bf or} $mediate \neq ${\bf false do} \nnl
\>\> {\bf return}; \nnl
\>$m'_i \leftarrow hasConflict(x_i)$; \nnl
\>{\bf if} \ $m'_i$ and $\neg \exists_j (p_j>p_i \land m_j =\,= {\bf true})$\nnl
\>\>{\bf if} $\exists(d'_i \in D_i)\ (d'_i \cup agent\_view$ does not conflict) \nnl
\>\>\ \ \ {\bf and} $d_i$ conflicts exclusively with lower priority neighbors \nnl
\>\>\> $d_i \leftarrow d'_i$; \nnl
\>\>\> send ({\bf ok?}, ($x_i, p_i, d_i, m_i$)) to all $x_j \in agent\_view$;\nnl
\>\>{\bf else} \nnl 
\>\>\>{\bf do mediate}; \nnl
\> {\bf else if} \ $m_i \neq m'_i$ \nnl
\>\> $m_i \leftarrow m'_i$; \nnl
\>\> send ({\bf ok?}, ($x_i, p_i, d_i, m_i$)) to all $x_j \in agent\_view$;\nnl
\> {\bf end if}; \nnl
{\bf end check\_agent\_view};
}
}
}
\caption{The procedures for doing local resolution, updating the {\it agent\_view} and the {\it good\_list}.}
\label{apo_ok}
\end{figure}

When an agent receives an initialization message (either during the
initialization or through a later link request), it records the
information in its $agent\_view$ and adds the variable to the
$good\_list$ if it can.  A variable is only added to the $good\_list$
if it is a neighbor of another variable already in the $good\_list$.  This
ensures that the graph created by the variables in the $good\_list$
always remains connected, which focuses the agent's internal problem
solving on variables which it knows it has an interdependency with.
The $initList$ is then checked to see if this message is a link
request or a response to a link request.  If an agent is in the
$initList$, it means that this message is a response, so the agent
removes the name from the \emph{initList} and does nothing further.
If the agent is not in the $initList$ then it means this is a request,
so a response ``init'' is generated and sent.

\begin{figure}
\centerline{
\hbox{
\progstyle{
procedure {\bf mediate} \nnl
\> $preferences\leftarrow \emptyset$; \nnl
\> $counter \leftarrow 0$; \nnl
\> {\bf for each} $x_j \in good\_list$ {\bf do} \nnl
\>\> send ({\bf evaluate?}, ($x_i, p_i$)) to $x_j$; \nnl
\>\> {\it counter} ++; \nnl
\>{\bf end do}; \nnl
\> {\it mediate} $\leftarrow$ {\bf true};\nnl
{\bf end mediate}; \nnl \nnl
{\bf when receive} ({\bf wait!}, ($x_j, p_j$)) {\bf do} \nnl
\> update $agent\_view$ with $(x_j, p_j)$; \nnl
\> {\it counter} -\,-; \nnl
\> {\bf if} {\it counter} == 0  {\bf do choose\_solution}; \nnl
{\bf end do};\nnl\nnl
{\bf when receive} ({\bf evaluate!}, ($x_j, p_j, labeled\ D_j$)) {\bf do}\nnl
\> record ($x_j, labeled\ D_j$) in {\it preferences}; \nnl
\> update $agent\_view$ with $(x_j, p_j)$; \nnl
\> {\it counter} -\,-; \nnl
\> {\bf if} {\it counter} == 0  {\bf do choose\_solution}; \nnl
{\bf end do};
}
}
}
\caption{The procedures for mediating an APO session.}
\label{apo_mediator}
\end{figure}

It is important to note that the agents contained in the $good\_list$
are a subset of the agents contained in the $agent\_view$.  This is
done to maintain the integrity of the $good\_list$ and allow links to
be bidirectional.  To understand this point, consider the case when a
single agent has repeatedly mediated and has extended its local
subproblem down a long path in the constraint graph.  As it does so,
it links with agents that may have a very limited view and therefore
are unaware of their indirect connection to the mediator.  In order
for the link to be bidirectional, the receiver of the link request has
to store the name of the requester in its $agent\_view$, but cannot
add them to their $good\_list$ until a path can be identified.  As can
be seen in section \ref{APO:proof}, the bi-directionality of links is
important to ensure the protocol's soundness.

\subsubsection{Checking the agent view (Figure \ref{apo_ok})}
\label{APO:checkview}
After all of the initialization messages are received, the agents
execute the check\_agent\_view procedure (at the end of figure \ref{apo_init}).  In this procedure, the
current $agent\_view$ (which contains the assigned, known variable
values) is checked to identify conflicts between the variable owned by
the agent and its {\bf neighbors}.  If, during this check (called {\it hasConflict} in the figure), an agent finds a
conflict with one or more of its neighbors and has not been told by a
higher priority agent that they want to mediate, it assumes the role
of the mediator.

\begin{figure*}
\centerline{
\hbox{
\progstyle{
procedure {\bf choose\_solution} \nnl
\>select a solution {\it s} using a Branch and Bound search that: \nnl
\>\>\ \ \ 1. satisfies the constraints between agents in the {\it good\_list} \nnl
\>\>\ \ \ 2. minimizes the violations for agents outside of the session \nnl
\> {\bf if} $\neg \exists s$ that satisfies the constraints {\bf do}\nnl
\>\> {\bf broadcast no solution}; \nnl
\> {\bf for each} $x_j \in agent\_view$ {\bf do} \nnl
\>\> {\bf if} $x_j \in preferences$ {\bf do} \nnl
\>\>\> {\bf if} $d'_j \in s$ violates an $x_k$ {\bf and} $x_k \notin agent\_view$ {\bf do} \nnl
\>\>\>\> send ({\bf init}, ($x_i, p_i, d_i, m_i, D_i, C_i$)) to $x_k$; \nnl
\>\>\>\> add $x_k$ to {\it initList}; \nnl
\>\>\> {\bf end if}; \nnl
\>\>\> send ({\bf accept!}, ($d'_j, x_i, p_i, d_i, m_i$)) to $x_j$; \nnl
\>\>\> update {\it agent\_view} for $x_j$ \nnl
\>\>{\bf else} \nnl
\>\>\> send ({\bf ok?}, ($x_i, p_i, d_i, m_i$)) to $x_j$; \nnl
\>\> {\bf end if}; \nnl
\> {\bf end do}; \nnl
\> {\it mediate} $\leftarrow$ {\bf false};\nnl
\> {\bf check\_agent\_view};\nnl
{\bf end choose\_solution};
}
}
}
\caption{The procedure for choosing a solution during an APO mediation.}
\label{apo_choose}
\end{figure*}

An agent can tell when a higher priority agent wants to mediate
because of the $m_i$ flag mentioned in the previous section.  Whenever
an agent checks its $agent\_view$ it recomputes the value of this flag
based on whether or not it has existing conflicts with its neighbors.
When this flag is set to {\bf true} it indicates that the agent wishes
to mediate if it is given the opportunity. This mechanism acts like a
two-phase commit protocol, commonly seen in database systems, and
ensures that the protocol is live-lock and dead-lock free.

When an agent becomes the mediator, it first attempts to rectify the
conflict(s) with its neighbors by changing its own variable.  This
simple, but effective technique prevents mediation sessions from
occurring unnecessarily, which stabilizes the system and saves messages
and time.  If the mediator finds a value that removes the conflict, it
makes the change and sends out an ``ok?'' message to the agents in its
$agent\_view$.  If it cannot find a non-conflicting value, it starts a
mediation session.  An ``ok?'' message is similar to an ``init''
message, in that it contains information about the priority, current
value, etc. of a variable.

\begin{figure}
\centerline{
\hbox{
\progstyle{
{\bf when received} ({\bf evaluate?}, $(x_j, p_j)$) {\bf do} \nnl
\> $m_j \leftarrow {\bf true}$; \nnl
\> {\bf if} {\it mediate} == {\bf true or} $\exists_k (p_k>p_j \land m_k =\,= {\bf true})\ {\bf do}$\nnl
\>\> send ({\bf wait!}, $(x_i, p_i)$); \nnl
\> {\bf else} \nnl
\>\> {\it mediate} $\leftarrow$ {\bf true};\nnl
\>\> label each $d \in D_i$ with the names of the agents \nnl
\>\>\ \ \ that would be violated by setting $d_i\leftarrow d$; \nnl
\>\> send ({\bf evaluate!}, ($x_i, p_i, labeled\ D_i$)); \nnl
\>{\bf end if}; \nnl
{\bf end do}; \nnl \nnl
{\bf when received} ({\bf accept!}, $(d, x_j, p_j, d_j, m_j)$) {\bf do} \nnl
\> $d_i \leftarrow d$; \nnl
\> {\it mediate} $\leftarrow$ {\bf false};\nnl
\> send ({\bf ok?}, ($x_i, p_i, d_i, m_i$)) to all $x_j$ in {\it agent\_view}; \nnl
\> update {\it agent\_view} with $(x_j, p_j, d_j, m_j)$; \nnl
\> {\bf check\_agent\_view}; \nnl
{\bf end do};
}
}
}
\caption{Procedures for receiving an APO session.}
\label{apo_receive}
\end{figure}

\subsubsection{Mediation (Figures \ref{apo_mediator}, \ref{apo_choose}, and \ref{apo_receive})}
The most complex and certainly most interesting part of the protocol
is the mediation.  As was previously mentioned in this section, an
agent decides to mediate if it is in conflict with one of its
neighbors and is not expecting a session request from a higher
priority agent.  The mediation starts with the mediator sending out
``evaluate?'' messages to each of the agents in its $good\_list$.  The
purpose of this message is two-fold.  First, it informs the receiving
agent that a mediation is about to begin and tries to obtain a lock
from that agent.  This lock, referred to as $mediate$ in the figures,
prevents the agent from engaging in two sessions simultaneously or
from doing a local value change during the course of a session.  The
second purpose of the message is to obtain information from the agent
about the effects of making them change their local value.  This is a
key point.  By obtaining this information, the mediator gains
information about variables and constraints outside of its local view
without having to directly and immediately link with those agents.
This allows the mediator to understand the greater impact of its
decision and is also used to determine how to extend its view once it
makes its final decision.

When an agent receives a mediation request, it responds with either a
``wait!'' or ``evaluate!'' message.  The ``wait'' message indicates to
the requester that the agent is currently involved in a session or is
expecting a request from an agent of higher priority than the
requester, which in fact could be itself.  If the agent is available,
it labels each of its domain elements with the names of the agents
that it would be in conflict with if it were asked to take that value.
This information is returned in the ``evaluate!'' message.

The size of the ``evaluate!'' message is strongly related to the
number of variables and the size of the agent's domain.  In cases
where either of these are extremely large, a number of techniques can
be used to reduce the overall size of this message.  Some example
techniques include standard message compression, limiting the domain
elements that are returned to be only ones that actually create
conflict or simply sending relevant value/variable pairs so the
mediator can actually do the labeling.  This fact means that the
largest ``evaluate!'' message ever actually needed is polynomial in
the number of agents ($O(V)$). In this implementation, for graph
coloring, the largest possible ``evaluate!'' message is $O(|D_i| +
|V|)$.

It should be noted that the agents do not need to return all of the
names when they have privacy or security reasons.  This effects the
completeness of the algorithm, because the completeness relies on one
or more of the agents eventually centralizing the entire problem in
the worst case.  As was mentioned in section \ref{motivation},
whenever an agent attempts to completely hide information about a
shared variable or constraint in a distributed problem, the
completeness is necessarily effected.
 
When the mediator has received either a ``wait!'' or ``evaluate!'' 
message from the agents that it sent a request to, it chooses a
solution.  The mediator determines that it has received all of the
responses by using the \emph{counter} variable which is set to the
size of the good\_list when the ``evaluate?'' messages are first sent.
As the mediator receives either a ``wait!'' or ``evaluate!'' message,
it decrements this counter.  When it reaches 0, all of the agents have
replied.

Agents that sent a "wait!" message are dropped from the mediation while
for agents that sent an "evaluate!" message their labeled domains
specified in the message are recorded and used in the search process.  The
mediator then uses the current values along with the labeled domains it has
received in the ``evaluate!'' messages to conduct a centralized search.

Currently, solutions are generated using a Branch and Bound search
\cite{freuder92} where all of the constraints in the good\_list must
be satisfied and the number of outside conflicts are minimized.  This is very similar to the the min-conflict heuristic \cite{minton92}.  Notice that although
this search takes all of the variables and constraints in its
good\_list into consideration, the solution it generates may not
adhere to the variable values of the agents that were dropped from the
session.  These variables are actually considered outside of the
session and the impact of not being able to change their values is
calculated as part of the min-conflict heuristic.  This causes the
search to consider the current values of dropped variables as
weak-constraints on the final solution.

In addition, the domain for each of the variables in the good\_list is
ordered such that the variable's current value is its first element.
This causes the search to use the current value assignments as the
first path in the search tree and has the tendency to minimize the
changes being made to the current assignments.  These heuristics, when
combined together, form a lock and key mechanism that simultaneously
exploits the work that was previously done by other mediators and acts
to minimize the number of changes in those assignments.  As will be
presented in section \ref{APO:evaluation}, these simple feed-forward
mechanisms, combined with the limited centralization needed to solve
problems, account for considerable improvements in the algorithms
runtime performance.

If no satisfying assignments are found during this search, the
mediator announces that the problem is unsatisfiable and the algorithm
terminates.  Once the solution is chosen, ``accept!'' messages are
sent to the agents in the session, who, in turn, adopt the proposed
answer.

The mediator also sends ``ok'' messages to the agents that are in its
$agent\_view$, but for whatever reason were not in the session.  This
simply keeps those agents' $agent\_view$s up-to-date, which is
important for determining if a solution has been reached.  Lastly,
using the information provided to it in the ``evaluate!'' messages,
the mediator sends ``init'' messages to any agent that is outside of
its $agent\_view$, but it caused conflict for by choosing a solution.
This ``linking'' step extends the mediators view along paths that are
likely to be critical to solving the problem or identifying an
over-constrained condition. This step also ensures the completeness of
the protocol.

Although termination detection is not explicitly part of the APO
protocol, a technique similar to \cite{wellman99distributed} could
easily be added to detect quiescence amongst the agents.

\subsubsection{Multiple Variables and $n$-ary Constraints}
\label{APO:remove}
Removing the restrictions presented in section \ref{APO:dcsp} is a
fairly straightforward process.  Because APO uses linking as part of
the problem solving process, working with $n$-ary constraints simply
involves linking with the $n$ agents within the constraint during
initialization and when a post-mediation linking needs to occur.
Priorities in this scheme are identical to those used for binary
constraints.

Removing the single agent per variable restriction is also not very
difficult and in fact is one of the strengths of this approach.  By
using a spanning tree algorithm on initialization, the agents can
quickly identify the interdependencies between their internal
variables which they can then use to create separate $good\_list$s for
each of the disconnected components of their internal constraint
graph.  In essence, on startup, the agents would treat each of these
decomposed problems as a separate problem, using a separate $m_i$
flag, priority, $good\_list$, etc.  As the problem solving unfolds,
and the agent discovers connections between the internal variables
(through external constraints), these decomposed problems could be
merged together and utilize a single structure for all of this
information.

This technique has the advantages of being able to ensure consistency
between dependent internal variables before attempting to mediate
(because of the local checking before mediation), but allows the agent
to handle independent variables as separate problems.  Using a situation aware technique such as this one has been shown to yield the best results in previous work\cite{Mammen-73}.  In addition, this technique allows the agents to hide
variables that are strictly internal.  By doing pre-computation on the
decomposed problems, the agents can construct constraints which
encapsulate each of the subproblems as n-ary constraints where n is the
number of variables that have external links.  These derived
constraints can then be sent as part of the ``init'' message whenever
the agent receives a link request for one of its external variables.
 
\begin{figure*}
  \epsfxsize=4.0in 
  \hspace*{\fill} 
  \epsffile{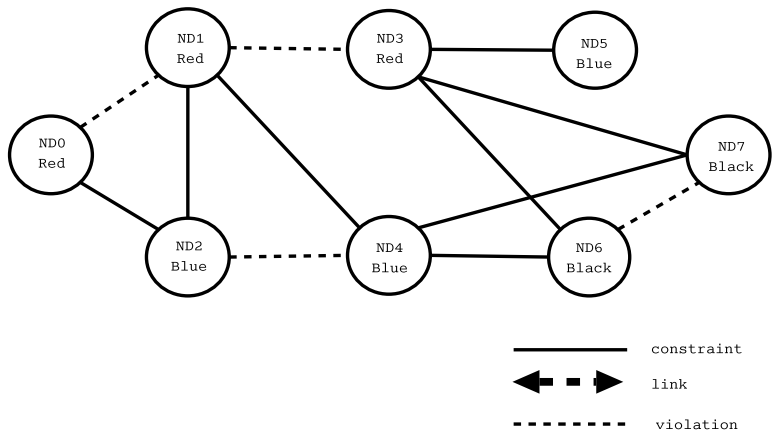}
  \hspace*{\fill} 
  \caption{An example of a 3-coloring problem with 8 nodes and 12 edges.}  
  \label{APO:base-ex}
\end{figure*}

\subsection{An Example}  
\label{APO:example}
Consider the 3-coloring problem presented in figure \ref{APO:base-ex}.
In this problem, there are 8 agents, each with a variable and 12 edges
or constraints between them.  Because this is a 3-coloring problem,
each variable can only be assigned one of the three available colors
\{Black, Red, or Blue\}.  The goal is to find an assignment of colors
to the variables such that no two variables, connected by an edge,
have the same color.

In this example, four constraints are in violation: (ND0,ND1),
(ND1,ND3), (ND2,ND4), and (ND6,ND7).  Following the algorithm, upon
startup each agent adds itself to its $good\_list$ and sends an
``init'' message to its neighbors.  Upon receiving these messages, the
agents add each of their neighbors to their $good\_list$ because they
are able to identify a shared constraint with themselves.

Once the startup has been completed, each of the agents checks its
$agent\_view$.  All of the agents, except ND5, find that they have
conflicts.  ND0 (priority 3) waits for ND1 to mediate (priority 5).
ND6 and ND7, both priority 4, wait for ND4 (priority 5, tie with ND3
broken using lexicographical ordering).  ND1, having an equal number of
agents in its $good\_list$, but a lower lexicographical order, waits
for ND4 to start a mediation.  ND3, knowing it is highest priority
amongst its neighbors, first checks to see if it can resolve its
conflict by changing its value, which in this case, it cannot.  ND3
starts a session that involves ND1, ND5, ND6, and ND7.  It sends each
of them an ``evaluate?'' message.  ND4 being highest priority amongst
its neighbors, and unable to resolve its conflict locally, also starts
a session by sending ``evaluate?'' messages to ND1, ND2, ND6, and ND7.

When each of the agents in the mediation receives the ``evaluate?'' 
message, they first check to see if they are expecting a mediation
from a higher priority agent.  In this case, ND1, ND6, and ND7 are
expecting from ND4 so tell ND3 to wait.  Then they label their domain
elements with the names of the variables that they would be in
conflict with as a result of adopting that value.  This information is
sent in an ``evaluate!'' message.  The following are the labeled
domains for the agents that are sent to ND4:

\begin{itemize}
\item ND1 - Black causes no conflicts; Red conflicts with ND0 and ND3; Blue conflicts with ND2 and ND4
\item ND2 - Black causes no conflicts; Red conflicts with ND0 and ND3; Blue conflicts with ND4
\item ND6 - Black conflicts with ND7; Red conflicts with ND3; Blue conflicts with ND4
\item ND7 - Black conflicts with ND6; Red conflicts with ND3; Blue conflicts with ND4
\end{itemize}

\pagebreak

The following are the responses sent to ND3:

\begin{itemize}
\item ND1 - wait!
\item ND5 - Black causes no conflicts; Red conflicts with ND3; Blue causes no conflicts
\item ND6 - wait!
\item ND7 - wait!
\end{itemize}

\begin{figure*}
  \epsfxsize=4.0in \hspace*{\fill} \epsffile{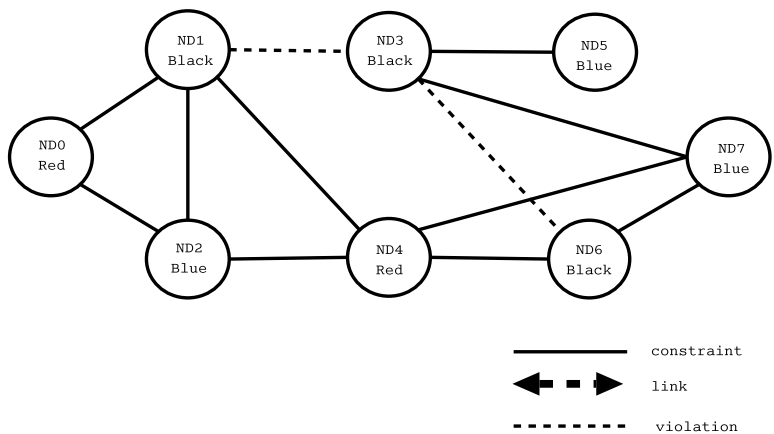}
  \hspace*{\fill} \caption{The state of the sample problem after ND3
  leads the first mediation.}  \label{APO:400-ex}
\end{figure*}

Once all of the responses are received, the mediators, ND3 and ND4,
conduct a branch and bound searches that attempt to find a satisfying
assignment to their subproblems and minimizes the amount of conflict that
would be created outside of the mediation.  If either of them cannot
find at least one satisfying assignment, it broadcasts that a solution
cannot be found.  

In this example, ND3, with the limited information that it has,
computes a satisfying solution that changes its own color and to
remain consistent would have also changed the colors of ND6 and ND7.
Since it was told by ND6 and ND7 to wait, it changes its color, sends
an ``accept'!'' message to ND5 and ``ok?'' messages to ND1, ND6 and
ND7.  Having more information, ND4 finds a solution that it thinks
will solve its subproblem without creating outside conflicts.  It
changes its own color to red, ND7 to blue, and ND1 to black leaving
the problem in the state shown in figure \ref{APO:400-ex}.

ND1, ND4, ND5, ND6 and ND7 inform the agents in their $agent\_view$ of
their new values, then check for conflicts.  This time, ND1, ND3, and ND6
notice that their values are in conflict.  ND3, having the highest
priority, becomes the mediator and mediates a session with ND1, ND5,
ND6, and ND7.  Following the protocol, ND3 sends out the ``evaluate?'' 
messages and the receiving agents label and respond.  The following
are the labeled domains that are returned:

\begin{figure*}
  \epsfxsize=4.0in 
  \hspace*{\fill}
  \epsffile{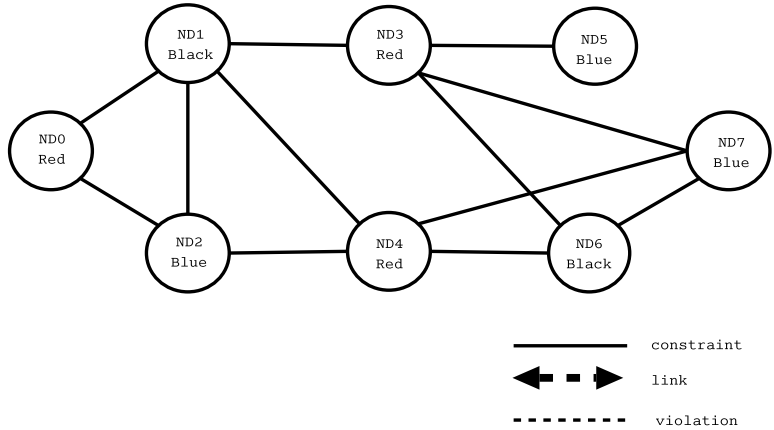} 
  \hspace*{\fill} 
  \caption{The final solution after ND2 leads the second mediation.}
  \label{APO:final-ex}
\end{figure*}

\begin{itemize}
\item ND1 - Black conflicts with ND3; Red conflicts with ND0 and ND4; Blue conflicts with ND2
\item ND5 - Black conflicts with ND3; Red causes no conflicts; Blue causes no conflicts
\item ND6 - Black conflicts with ND3; Red conflicts with ND4; Blue conflicts with ND7
\item ND7 - Black conflicts with ND3 and ND6; Red conflicts with ND4; Blue causes no conflicts
\end{itemize} 

ND3, after receiving these messages, conducts its search and finds a
solution that solves its subproblem.  It chooses to change its color
to red.  ND1, ND3, ND5, ND6, and ND7 check their $agent\_view$ and
find no conflicts.  Since, at this point, none of the agents have any
conflict, the problem is solved (see figure \ref{APO:final-ex}).

\subsection{Soundness and Completeness}
\label{APO:proof}

\newtheorem{lemma}{Lemma}
\newtheorem{theorem}{Theorem}
In this section we will show that the APO algorithm is both sound and
complete.  For these proofs, it is assumed that all communications are
reliable, meaning that if a message is sent from $x_i$ to $x_j$ that
$x_j$ will receive the message in a finite amount of time.  We also
assume that if $x_i$ sends message $m_1$ and then sends message $m_2$
to $x_j$, that $m_1$ will be received before $m_2$.  Lastly, we assume
that the centralized solver used by the algorithm is sound and
complete.  Before we prove the soundness and completeness, it helps to
have a few principal lemmas established.

\begin{lemma}
Links are bidirectional. i.e. If $x_i$ has $x_j$ in its $agent\_view$ then eventually $x_j$ will have $x_i$ in its $agent\_view$.
\label{APO:bidirectional}
\end{lemma}
{\bf Proof:}

Assume that $x_i$ has $x_j$ in its $agent\_view$ and that $x_i$ is not
in the $agent\_view$ of $x_j$.  In order for $x_i$ to have $x_j$ in
its $agent\_view$, $x_i$ must have received an ``init'' message at
some point from $x_j$.  There are two cases.

{\bf Case 1:}   $x_j$ is in the $initList$ of $x_i$. In this case, $x_i$ must have sent $x_j$ an ``init'' message first, meaning that $x_j$ received an ``init'' message and therefore has $x_i$ in its $agent\_view$, a contradiction.  

{\bf Case 2:}  $x_j$ is not in the $initList$ of $x_i$.  In this case, when $x_i$ receives the ``init'' message from $x_j$, it responds with an ``init'' message.  That means that if the reliable communication assumption holds, eventually $x_j$ will receive $x_i$'s ``init'' message and add $x_i$ to its $agent\_view$.  Also a contradiction.

\begin{lemma}
If agent $x_i$ is linked to $x_j$ and $x_j$ changes its value, then $x_i$ will eventually be informed of this change and update its agent\_view. 
\label{APO:uptodate}
\end{lemma}
{\bf Proof:}

Assume that $x_i$ has a value in its $agent\_view$ for $x_j$ that is incorrect.  This would mean that at some point $x_j$ altered its value without informing $x_i$.  There are two cases:

{\bf Case 1:}  $x_j$ did not know it needed to send $x_i$ an update.  i.e. $x_i$ was not in $x_j$'s $agent\_view$.  Contradicts lemma \ref{APO:bidirectional}.

{\bf Case 2:}  $x_j$ did not inform all of the agents in its $agent\_view$ when it changes its value.  It is clear from the code that this cannot happen.  Agents only change their values in the check\_agent\_view, choose\_solution, and accept! procedures.  In each of these cases, it informs all of the agents within its $agent\_view$ by either sending an ``ok?'' or through the ``accept!'' message that a change to its value has occurred.  A contradiction.

\begin{lemma}
If $x_i$ is in conflict with one or more of its neighbors, does not expect a mediation from another higher priority agent in its $agent\_view$, and is currently not in a session, then it will act as mediator.
\label{APO:willmediate}
\end{lemma}
{\bf Proof:}

Directly from the procedure check\_agent\_view.

\begin{lemma}
If $x_i$ mediates a session that has a solution, then each of the
constraints between the agents involved in the mediation will be
satisfied.
\label{APO:mediatefix}
\end{lemma}
{\bf Proof:}

Assume that there are two agents $x_j$ and $x_k$ (either of them could
be $x_i$), that were mediated over by $x_i$ and after the mediation
there is a conflict between $x_j$ and $x_k$.  There are two ways this
could have happened.

{\bf Case 1:} One or both of the agents must have a value that $x_i$
did not assign to it as part of the mediation.

Assume that $x_j$ and/or $x_k$ has a value that $x_i$ did not assign.
We know that since $x_i$ mediated a session including $x_j$ and $x_k$,
that $x_i$ did not receive a ``wait!'' message from either of $x_j$ or
$x_k$.  This means that they could not have been mediating.  This also
means that they must have set their $mediate$ flags to true when $x_i$
sent them the ``evaluate?'' message.  Since the only times an agent
can change its value is when its $mediate$ flag is false, it is
mediating, or has been told to by a mediator, $x_j$ and/or $x_k$ could
only have changed their values is if $x_i$ told them to, which
contradicts the assumption.

{\bf Case 2:} $x_i$ assigned them a value that caused them to be in
conflict with one another.

Let's assume that $x_i$ assigned them conflicting values.  This means
that $x_i$ chose a solution that did not take into account the
constraints between $x_j$ and $x_k$.  But, we know that $x_i$ only
chooses satisfying solutions that include all of the constraints
between all of the agents in the $good\_list$.  This leads to a
contradiction.

This lemma is important because it says that once a mediator has
successfully concluded its session, the only conflicts that can exist
are on constraints that were outside of the mediation.  This can be
viewed as the mediator pushing the constraint violations outside of
its view.  In addition, because mediators get information about who
the violations are being pushed to and establish links with those
agents, over time, they gain more context.  This is a very important
point when considering the completeness of the algorithm.

\begin{theorem}
The APO algorithm is sound.  i.e.  It reaches a stable state only if it has either found an answer or no solution exists.
\label{APO:soundness}
\end{theorem}
{\bf Proof:}

In order to be sound, the agents can only stop when they have reached
an answer.  The only condition in which they would stop without having
found an answer is if one or more of the agents is expecting a
mediation request from a higher priority agent that does not send it.
In other words, the protocol has deadlocked.

Let's say we have 3 agents, $x_i$, $x_j$, and $x_k$ with $p_i<p_j \vee p_k<p_j$ ($i$ could be equal to $k$) and $x_k$ has a conflict with $x_j$.  There are two cases in which $x_j$ would not mediate a session that included $x_i$, when $x_i$ was expecting it to:

{\bf Case 1:} $x_i$ has $m_j = {\bf true}$ in its $agent\_view$ when
the actual value should be {\bf false}.

Assume that $x_i$ has $m_j = {\bf true}$ in its $agent\_view$ when the
true value of $m_j = {\bf false}$.  This would mean that at some point
$x_j$ changed the value of $m_j$ to false without informing $x_i$.
There is only one place that $x_j$ changes the value of $m_j$, this is
in the check\_agent\_view procedure (see figure \ref{apo_ok}).  Note
that in this procedure, whenever the flag changes value from true to
false, the agent sends an ``ok'' message to all the agents in its
$agent\_view$.  Since by lemma \ref{APO:bidirectional} we know that
$x_i$ is in the $agent\_view$ of $x_j$, $x_i$ must have received the
message saying that $m_j = {\bf false}$, contradicting the assumption.

{\bf Case 2:} $x_j$ believes that $x_i$ should be mediating when $x_i$
does not believe it should be.  i.e. $x_j$ thinks that $m_i = {\bf
  true}$ and $p_i > p_j$.

By the previous case, we know that if $x_j$ believes that $m_i = {\bf
  true}$ that this must be the case.  We only need to show that $p_i <
p_j$.  Let's say that $p'_i$ is the priority that $x_j$ believes $x_i$
has and assume that $x_j$ believes that $p'_i > p_j$ when, in fact
$p_i < p_j$.  This means that at some point $x_i$ sent a message to
$x_j$ informing it that its current priority was $p'_i$.  Since we
know that priorities only increase over time (the good\_list only gets
larger), we know that $p'_i \leq p_i$ ($x_j$ always has the correct
value or underestimates the priority of $x_i$).  Since $p_j > p_i$ and
$p_i \geq p'_i$ then $p_j > p'_i$ which contradicts the assumption.

This is also an important point when considering how the algorithm
behaves.  This proof says that agents always either know or
underestimate the true value of their neighbors' priorities.  Because
of this, the agents will attempt to mediate when in fact sometimes,
they shouldn't.  The side effect of this attempt, however, is that the
correct priorities are exchanged so the same mistake doesn't get
repeated.  The other important thing to mention is the case were the
priority values become equal.  In this case, the tie is broken by
using the alphabetical order of names of the agents.  This ensures
that there is always a way to break ties.

\begin{defn}
Oscillation is a condition that occurs when a subset $V' \subseteq V$ of the agents are infinitely cycling through their allowable values without reaching a solution.  In other words, the agents are live-locked
\label{APO:oscillation}
\end{defn}

By this definition, in order to be considered part of an oscillation,
an agent within the subset must be changing its value (if it's stable,
it's not oscillating) and it must be connected to the other members of
the subset by a constraint (otherwise, it is not actually a part of
the oscillation).

\begin{theorem}
The APO algorithm is complete.  i.e. if a solution exists, the algorithm will find it.  If a solution does not exist, it it will report that fact.
\label{APO:completeness}
\end{theorem}
{\bf Proof:}  

A solution does not exist whenever the problem is over-constrained.
If the problem is over-constrained, the algorithm will eventually
produce a $good\_list$ where the variables within it and their
associated constraints lead to no solution.  Since a subset of the
variables is unsatisfiable, the entire problem is unsatisfiable,
therefore, no solution is possible.  The algorithm terminates with
failure if and only if this condition is reached.

Since we have now shown in Theorem \ref{APO:soundness} that whenever
the algorithm reaches a stable state, the problem is solved and that
when it finds a subset of the variables that is unsatisfiable it
terminates, we only need to show that it reaches one of these two
states in finite time.  The only way for the agents to not reach a
stable state is when one or more of the agents in the system is in an
oscillation.

There are two cases to consider, the easy case is when a single agent is oscillating ($|V'| = 1$) and the other case is when more than one agent is oscillating ($|V'| > 1$).

{\bf Case 1:}  There is an agent $x_i$ that is caught in an infinite loop and all other agents are stable.  

Let's assume that $x_i$ is in an infinite processing loop.  That means
that no matter what it changes its value to, it is in conflict with
one of its neighbors, because if it changed its value to something
that doesn't conflict with its neighbors, it would have a solution and
stop.  If it changes its value to be in conflict with some $x_j$ that
is higher priority than it, then $x_j$ will mediate with $x_i$,
contradicting the assumption that all other agents are stable.  If
$x_i$ changes its value to be in conflict with a lower priority agent,
then by lemma \ref{APO:willmediate}, it will act as mediator with its
neighbors.  Since it was assumed that each of the other agents is in a
stable state, then all of the agents in $x_i$'s $good\_list$ will
participate in the session and by lemma \ref{APO:mediatefix}, agent
$x_i$ will have all of its conflicts removed.  This means that $x_i$
will be in a stable state contradicting the assumption that it was in
an infinite loop.

{\bf Case 2:}  Two or more agents are in an oscillation.

Let's say we have a set of agents $V' \subseteq V$ that are in an
oscillation.  Now consider an agent $x_i$ that is within $V'$.  We
know that the only conditions in which $x_i$ changes its value is when
it can do so and solve all of its conflicts (a contradiction because
$x_i$ wouldn't be considered part of the oscillation), as the
mediator, or as the receiver of a mediation from some other agent in
$V'$.  The interesting case is when an agent acts as the mediator.

Consider the case when $x_i$ is the mediator and call the set of
agents that it is mediating over $V_i$.  We know according to
definition \ref{APO:oscillation} that after the mediation, that at
least one conflict must be created or remain otherwise the oscillation
would stop and the problem would be solved.  In fact, we know that
each of the remaining conflicts must contain an agent from the set
$V'-V_i$ by lemma \ref{APO:mediatefix}.  We also know that for each
violated constraint that has a member from $V_i$, that $x_i$ will link
with any agent that is part of those constraints and not a member of
$V_i$.  The next time $x_i$ mediates, the set $V_i$ will include these
members and the number of agents in the set $V'-V_i$ is reduced.  In
fact, whenever $x_i$ mediates the set $V'-V_i$ will be reduced
(assuming it is not told to wait! by one or more agents.  In this
case, it takes longer to reduce this set, but the proof still holds).
Eventually, after $O(|V|^2)$ mediations, some $x_i$ within $V'$ must
have $V_i = V'$ (every agent within the set must have mediated $|V'|$
times in order for this to happen).  When this agent mediates it will
push the violations outside of the set $V'$ or it will solve the
subproblem by lemma \ref{APO:mediatefix}. Either of these conditions
contradicts the oscillation assumption.  Therefore, the algorithm is
complete. {\bf QED}

It should be fairly clear that, in domains that are exponential, the
algorithm's worse-case runtime is exponential.  The space complexity
of the algorithm is, however, polynomial, because the agents only
retain the names, priorities, values, and constraints of other agents.

\section{Evaluation}
\label{APO:evaluation}
A great deal of testing and evaluation has been conducted on the APO
algorithm.  Almost exclusively, these test are done comparing the APO
algorithm with the currently fastest known, complete algorithm for
solving DCSPs called the Asynchronous Weak Commitment (AWC)
protocol. In this section we will describe the AWC protocol (section
\ref{AWC}), then will describe the distributed 3-coloring domain and
present results from extensive testing done in this domain (section
\ref{coloring}).  This testing compares these two algorithms across a
variety of metrics, including the cycle time, number of messages, and
serial runtime.

Next, we will describe the tracking domain (section \ref{tracking})
and present results from testing in this domain as well.  For this
domain, we modified the core search algorithm of APO to take advantage
of the polynomial complexity of this problem.  This variant called,
\emph{APO-Flow}, will also be described.

\subsection{The Asynchronous Weak Commitment (AWC) Protocol}
\label{AWC}
The AWC protocol \cite{yokoo95awc} was one of the first algorithms
used for solving DCSPs.  Like the APO algorithm, AWC is based on
variable decomposition.  Also, like APO, AWC assigns each agent a
priority value that dynamically changes.  AWC, however, uses the
weak-commitment heuristic \cite{yokoo94:wks} to assign these
priorities values which is where it gets its name.

Upon startup, each of the agents selects a value for its variable and
sends ``ok?'' messages to its neighbors (agents it shares a constraint
with).  This message includes the variables value and priority (they
all start at 0).

When an agent receives an ``ok?'' message, it updates its
$agent\_view$ and checks its \emph{nogood list} for violated
$nogoods$.  Each nogood is composed of a set of nogood pairs which
describe the combination of agents and values that lead to an
unsatisfiable condition.  Initially, the only nogoods in an agent's
nogood list are the constraints on its variable.

When checking its nogood list, agents only check for violations of
higher priority nogoods.  The priority of a nogood is defined as the
priority of the lowest priority variable in the nogood.  If this value
is greater than the priority of the agent's variable, the nogood is
higher priority.  Based on the results from this check, one of three
things can happen:

\begin{enumerate}

\item If no higher priority nogoods are violated, the agent does nothing.

\item If there are higher priority nogoods that are violated and this can be repaired by simply changing the agent's variable value, then the agent changes its value and sends out ``ok?'' messages to the agents in its $agent\_view$.  If there are multiple possible satisfying values, then the agent chooses the one that minimizes the number of violated lower priority nogoods. 

\item If there are violated higher priority nogoods and this cannot be repaired by changing the value of its variable, the agent generates a new nogood.  If this nogood is the same as a previously generated nogood, it does nothing.  Otherwise, it then sends this new nogood to every agent that has a variable contained in the nogood and raises the priority value of its variable.  Finally, it changes its variable value to one that causes the least amount of conflict and sends out ``ok?'' messages. 
\end{enumerate}

Upon receiving a nogood message from another agent, the agent adds the
nogood to its nogood list and rechecks for nogood violations.  If the
new nogood includes the names of agents that are not in its
$agent\_view$ it links to them.  This linking step is essential to the
completeness of the search\cite{yokoo92abt},
but causes the agents to communicate nogoods and ok? messages to
agents that are not their direct neighbors in the constraint graph.
The overall effect is an increase in messages and a reduction in the
amount of privacy being provided to the agents because they
communicate potential domain values and information about their
constraints through the exchange of ok? and nogood messages with a
larger number of agents.

One of the more recent advances to the AWC protocol has been the
addition of resolvent-based nogood learning \cite{hirayama00learning}
which is an adaptation of classical nogood learning methods
\cite{ginsberg93:dynamicbt,cha96adding,frost94dead}.

The resolvent method is used whenever an agent finds that it needs to
generate a new nogood. Agents only generate new nogoods when each of
their domain values are in violation with at least one higher priority
nogood already in their nogood list.  The resolvent method works by
selecting one of these higher priority nogoods for each of the domain
values and aggregating them together into a new nogood.  This is
almost identical to a \emph{resolvent} in propositional logic which is
why it is referred to as resolvent-based learning.  The AWC protocol
used for all of our testing incorporates resolvent-based nogood
learning.

\subsection{Distributed Graph Coloring}
\label{coloring}
Following directly from the definition for a CSP, a graph coloring
problem, also known as a $k$-colorability problem, consists of the
following:
\begin{itemize}
\item a set of {\it n} variables $V = \{x_1,\ldots,x_n\}$.
\item a set of possible colors for each of the variables $D = \{D_1,\ldots,D_n\}$ where each $D_i$ is has exactly $k$ allowable colors.
\item a set of constraints $R = \{R_1,\ldots,R_m\}$ where each $R_i(d_i,d_j)$ is predicate which implements the ``not equals'' relationship.  This predicate returns true iff the value assigned to $x_i$ differs from the value assigned to $x_j$.
\end{itemize}

\begin{figure*}
\begin{center}
   \epsfxsize=5.0in 
   \epsffile{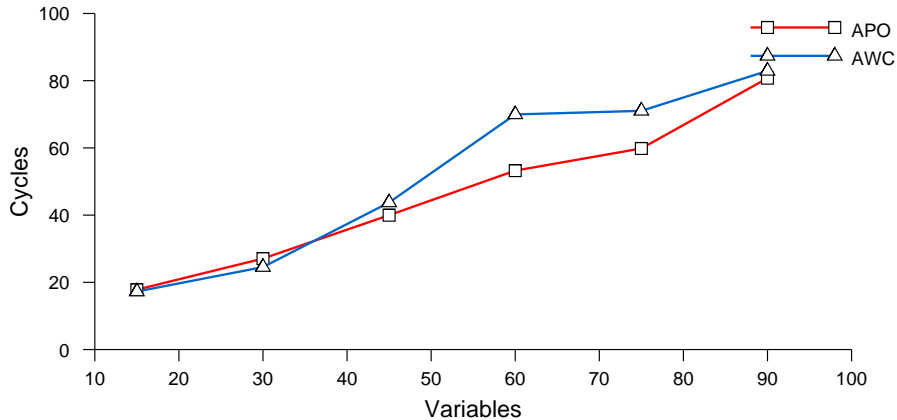}
  \caption{Comparison of the number of cycles needed to solve satisfiable, low-density 3-coloring problems of various sizes by AWC and APO.}
   \label{sattime2.0}
\end{center}
\end{figure*}

The problem is to find an assignment $A=\{d_1,\ldots,d_n|d_i \in
D_i\}$ such that each of the constraints in $R$ is satisfied.  Like
the general CSP, graph coloring has been shown to be NP-complete for
all values of $k>2$.

To test the APO algorithm, we implemented the AWC and APO algorithms
and conducted experiments in the distributed 3-coloring domain.  The
distributed 3-coloring problem is a 3-coloring problem with {\it n}
variables and {\it m} binary constraints where each agent is given a
single variable.  We conducted 3 sets of graph coloring based
experiments to compare the algorithms' computation and communication
costs.

\subsubsection{Satisfiable Graphs}
In the first set of experiments, we created solvable graph instances
with $m=2.0n$ ({\it low-density}), $m=2.3n$ ({\it medium-density}),
and $m=2.7n$ ({\it high-density}) according to the method presented in
\cite{minton92}.  Generating graphs in this way involves partitioning
the variables into $k$ equal-sized groups.  Edges are then added by
selecting two of the groups at random and adding an edge between a
random member of each group.  This method ensures that the resulting
graphs are satisfiable, but also tests a very limited and very likely
easier subset of the possible graphs.  These tests were done because
they are traditionally used by other researchers in DCSPs.

These particular values for $m$ were chosen because they represent the
three major regions within the phase-transition for 3-colorability
\cite{culberson01frozen}.  A phase transition in a CSP is defined
based on an order parameter, in this case the average node degree $d$.
The transition occurs at the point where random graphs created with
that order value yield half satisfiable and half unsatisfiable
instances.  Values of the order parameter that are lower than the
transition point (more than 50\% of the instance are satisfiable) are
referred to being to the left of the transition.  The opposite is
true of values to the right.

\begin{figure*}
\begin{center}
   \epsfxsize=5.0in 
   \epsffile{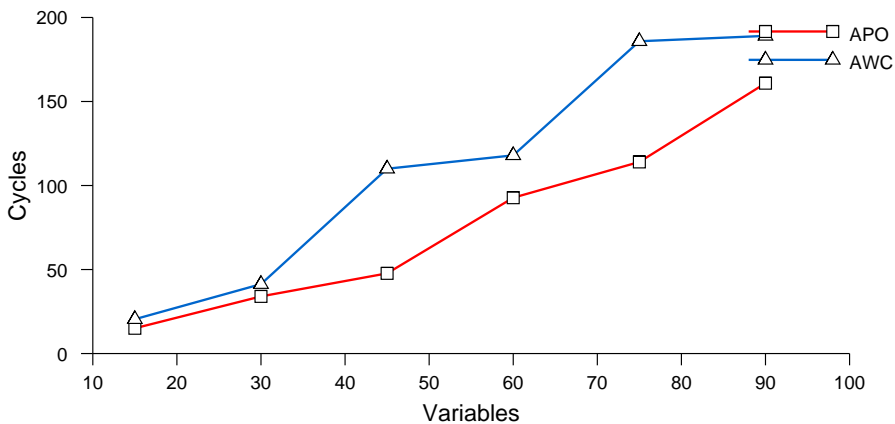}
  \caption{Comparison of the number of cycles needed to solve satisfiable, medium-density 3-coloring problems of various sizes by AWC and APO.}
   \label{sattime2.3}
\end{center}
\end{figure*}

Phase transitions are important because that are strongly correlated
with the overall difficulty of finding a solution to the graph
\cite{cheeseman91,monasson99,culberson01frozen}.  Within the phase
transition, randomly created instances are typically difficult to
solve.  Interestingly, problems to the right and left of the phase
transitions tend to be much easier.

In 3-colorability, the value $d = 2.0$ is to the left of the phase
transition.  In this region, randomly created graphs are very likely
to be satisfiable and are usually easy to find solve.  At
$d=2.3$, which is in the middle of the phase transition the graph has
about a 50\% chance of being satisfiable and is usually hard to solve.
For $m=2.7n$, right of the phase transition, graphs are more than
likely to be unsatisfiable and, again, are also easier to solve.

\begin{figure*}
\begin{center}
   \epsfxsize=5.0in 
   \epsffile{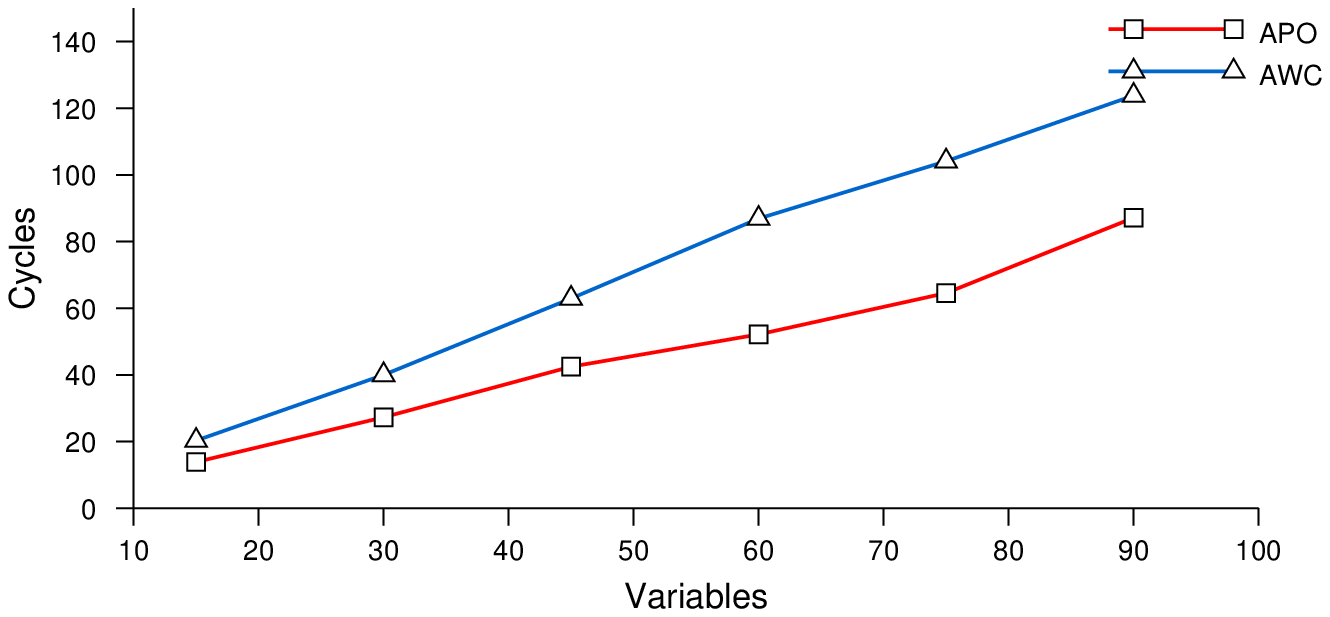}
  \caption{Comparison of the number of cycles needed to solve satisfiable, high-density 3-coloring problems of various sizes by AWC and APO.}
   \label{sattime2.7}
\end{center}
\end{figure*}

\begin{table*}
\begin{center}

   \begin{tabular}{|c|c|r|r|r|r|r|}
      \hline
       & & \multicolumn{1}{c|}{APO}&  \multicolumn{1}{c|}{APO} &  \multicolumn{1}{c|}{AWC} & \multicolumn{1}{c|}{AWC} &\\ 
        $d$  & Nodes & \multicolumn{1}{c|}{Mean} &  \multicolumn{1}{c|}{StDev} & \multicolumn{1}{c|}{Mean} & \multicolumn{1}{c|}{StDev} & p($AWC \leq APO$)\\
      \hline
            & 15 & 17.82 & 8.15 & 17.38 & 12.70 & 0.77 \\ 
            & 30 & 27.07 & 17.11 & 24.62 & 19.23 & 0.27 \\
        2.0 & 45 & 39.97 & 25.79 & 43.76 & 30.98 & 0.34 \\
            & 60 & 53.24 & 32.32 & 69.96 & 49.03 & 0.01 \\
            & 75 & 59.83 & 35.35 & 80.32 & 103.76 & 0.07 \\
            & 90 & 80.75 & 54.30 & 82.92 & 61.01 & 0.81 \\
            \cline{2-7}
	    & Overall &  &  &  &  & 0.01\\
      \hline
            & 15 & 15.04 & 5.55 & 20.49 & 11.27 & 0.00 \\ 
            & 30 & 34.01 & 16.81 & 41.30 & 29.58 & 0.04 \\
        2.3 & 45 & 47.72 & 26.58 & 109.99 & 74.85 & 0.00 \\
            & 60 & 92.73 & 72.46 & 135.60 & 146.57 & 0.01 \\
            & 75 & 114.02 & 75.84 & 185.84 & 119.94 & 0.00 \\
            & 90 & 160.88 & 125.12 & 189.04 & 91.27 & 0.07 \\
            \cline{2-7}
	    & Overall &  &  &  &  & 0.00\\
      \hline
	    & 15 & 13.83 & 3.56 & 20.29 & 10.32 & 0.00 \\ 
            & 30 & 27.28 & 10.10 & 39.99 & 24.08 & 0.00 \\
        2.7 & 45 & 42.47 & 18.01 & 62.92 & 35.23 & 0.00 \\
            & 60 & 52.15 & 23.12 & 86.89 & 43.69 & 0.00 \\
            & 75 & 64.54 & 26.26 & 104.09 & 46.62 & 0.00 \\
            & 90 & 87.14 & 42.82 & 127.04 & 64.97 & 0.00 \\
            \cline{2-7}
	    & Overall &  &  &  &  & 0.00\\
      \hline
   \end{tabular}
\end{center}
   \caption{Comparison of the number of cycles needed to solve satisfiable 3-coloring problems of various sizes and densities by AWC and APO.}
   \label{table-satcyc}
\end{table*}

A number of papers
\cite{yokoo00review,yokoo96dbo,hirayama00learning} have reported that
$m=2.7n$ is within the critical phase transition for 3-colorability.
This seems to have been caused by a misinterpretation of previous work in this area\cite{cheeseman91}.  Although Cheeseman, Kanefsky, and
Taylor reported that $m=2.7n$ was within the critical region for
3-colorability, they were using \emph{reduced} graphs for their
analysis.

A reduced graph is one in which the trivially colorable nodes and
non-relevant edges have been removed.  For example, one can easily
remove any node with just two edges in a 3-coloring problem because it
can always be trivially colored.  Additionally, nodes that possess a
unique domain element from their neighbors can also be easily removed.

In later work, Culberson and Gent identified the critical region as
being approximately $d=2.3$ and therefore was included in our tests\cite{culberson01frozen}.
One should note, however, that because phase transitions are typically
done on completely random graphs and by its very definition involves
both satisfiable and unsatisfiable instances, it is very hard to apply
the phase-tranisition results to graphs created using the
technique described at the beginning of the section because it only generates
satisfiable instances.  A detailed phase transition analysis has not
been done on this graph generation technique and in fact, we believe
that these graphs tend to be easier than randomly created satisfiable
ones of the same size and order.

To evaluate the relative strengths and weakness of each of the
approaches, we measured the number of {\it cycles} and the number of
messages used during the course of solving each of the problems.
During a cycle, incoming messages are delivered, the agent is allowed
to process the information, and any messages that were created during
the processing are added to the outgoing queue to be delivered at the
beginning of the next cycle.  The actual execution time given to one
agent during a cycle varies according to the amount of work needed to
process all of the incoming messages.  The random seeds used to create
each graph instance and variable instantiation were saved and used by
each of the algorithms for fairness.

\begin{table*}[tp]
\begin{center}
   \begin{tabular}{|c|c|r|r|r|r|}
      \hline
       && \multicolumn{1}{c|}{\% Links}&  \multicolumn{1}{c|}{\% Links} &  \multicolumn{1}{c|}{\% Central} & \multicolumn{1}{c|}{\% Central} \\ 
       & Nodes & \multicolumn{1}{c|}{Mean} &  \multicolumn{1}{c|}{StDev} & \multicolumn{1}{c|}{Mean} & \multicolumn{1}{c|}{StDev} \\
      \hline
      & 15 & 32.93 & 3.52 & 60.53 & 9.12 \\  
      & 30 & 17.24 & 2.59 & 42.53 & 8.06 \\ 
      APO & 45 & 11.97 & 1.84 & 33.27 & 6.61 \\ 
      & 60 & 9.30 & 1.30 & 29.00 & 7.31 \\ 
      & 75 & 7.42 & 0.95 & 24.60 & 6.73 \\ 
      & 90 & 6.51 & 0.96 & 24.93 & 6.16 \\ 
      \hline
      & 15 & 55.76 & 12.31 & 79.73 & 11.45 \\ 
      & 30 & 32.89 & 8.88 & 60.97 & 10.98 \\ 
      AWC & 45 & 27.00 & 7.89 & 56.49 & 11.57 \\ 
      & 60 & 26.47 & 9.12 & 56.98 & 14.23 \\ 
      & 75 & 21.99 & 7.67 & 53.19 & 13.29 \\ 
      & 90 & 20.11 & 7.89 & 50.38 & 13.78 \\ 
      \hline
   \end{tabular}
\end{center}

   \caption{Link statistics for satisfiable, low-density problems.}
   \label{table-satlinks2.0}
\end{table*}

\begin{table*}[tp]
\begin{center}
   \begin{tabular}{|c|c|r|r|r|r|}
      \hline
       && \multicolumn{1}{c|}{\% Links}&  \multicolumn{1}{c|}{\% Links} &  \multicolumn{1}{c|}{\% Central} & \multicolumn{1}{c|}{\% Central} \\ 
       &Nodes & \multicolumn{1}{c|}{Mean} &  \multicolumn{1}{c|}{StDev} & \multicolumn{1}{c|}{Mean} & \multicolumn{1}{c|}{StDev} \\
      \hline
      & 15 & 37.46 & 3.45 & 67.60 & 8.26\\ 
      & 30 & 21.24 & 2.85 & 51.37 & 9.32 \\ 
      APO & 45 & 14.90 & 2.27 & 43.67 & 11.40 \\ 
      & 60 & 12.85 & 2.99 & 42.98 & 13.06 \\ 
      & 75 & 11.17 & 2.45 & 43.76 & 12.69 \\ 
      & 90 & 10.07 & 3.10 & 40.47 & 14.55 \\ 
      \hline
      & 15 & 63.19 & 12.06 & 83.56 & 8.96 \\ 
      & 30 & 42.84 & 11.83 & 72.20 & 12.09 \\ 
      AWC & 45 & 52.05 & 13.52 & 80.67 & 11.24 \\
      & 60 & 43.69 & 14.59 & 74.93 & 14.56 \\  
      & 75 & 47.77 & 14.10 & 79.41 & 12.18 \\ 
      & 90 & 44.04 & 10.84 & 77.98 & 10.52 \\ 
      \hline
   \end{tabular}
   \end{center}
   \caption{Link statistics for satisfiable, medium-density problems.}
   \label{table-satlinks2.3}

\end{table*}

\begin{table*}[tp]
\begin{center}
   \begin{tabular}{|c|c|r|r|r|r|}
      \hline
       && \multicolumn{1}{c|}{\% Links}&  \multicolumn{1}{c|}{\% Links} &  \multicolumn{1}{c|}{\% Central} & \multicolumn{1}{c|}{\% Central} \\ 
       &Nodes & \multicolumn{1}{c|}{Mean} &  \multicolumn{1}{c|}{StDev} & \multicolumn{1}{c|}{Mean} & \multicolumn{1}{c|}{StDev} \\
      \hline
      & 15 & 43.28 & 2.86 & 74.53 & 8.34 \\ 
      & 30 & 24.29 & 2.61 & 59.30 & 9.75 \\
      APO & 45 & 18.07 & 2.06 & 52.62 & 10.21 \\
      & 60 & 13.86 & 1.77 & 47.78 & 11.32 \\
      & 75 & 11.85 & 1.57 & 46.37 & 11.32 \\
      & 90 & 10.86 & 1.85 & 50.73 & 14.48 \\
      \hline
      & 15 & 78.87 & 12.29 & 93.60 & 7.75 \\ 
      & 30 & 58.03 & 11.66 & 86.27 & 9.54 \\
      AWC & 45 & 54.06 & 13.44 & 82.67 & 13.34 \\
      & 60 & 53.01 & 12.77 & 83.47 & 11.04 \\
      & 75 & 49.63 & 11.36 & 81.91 & 9.93 \\
      & 90 & 47.72 & 13.81 & 80.32 & 15.06 \\
      \hline
   \end{tabular}
\end{center}
   \caption{Link statistics for satisfiable, high-density problems.}
   \label{table-satlinks2.7}
\end{table*}

\begin{figure*}[t]
\begin{center}
   \epsfxsize=5.0in 
   \epsffile{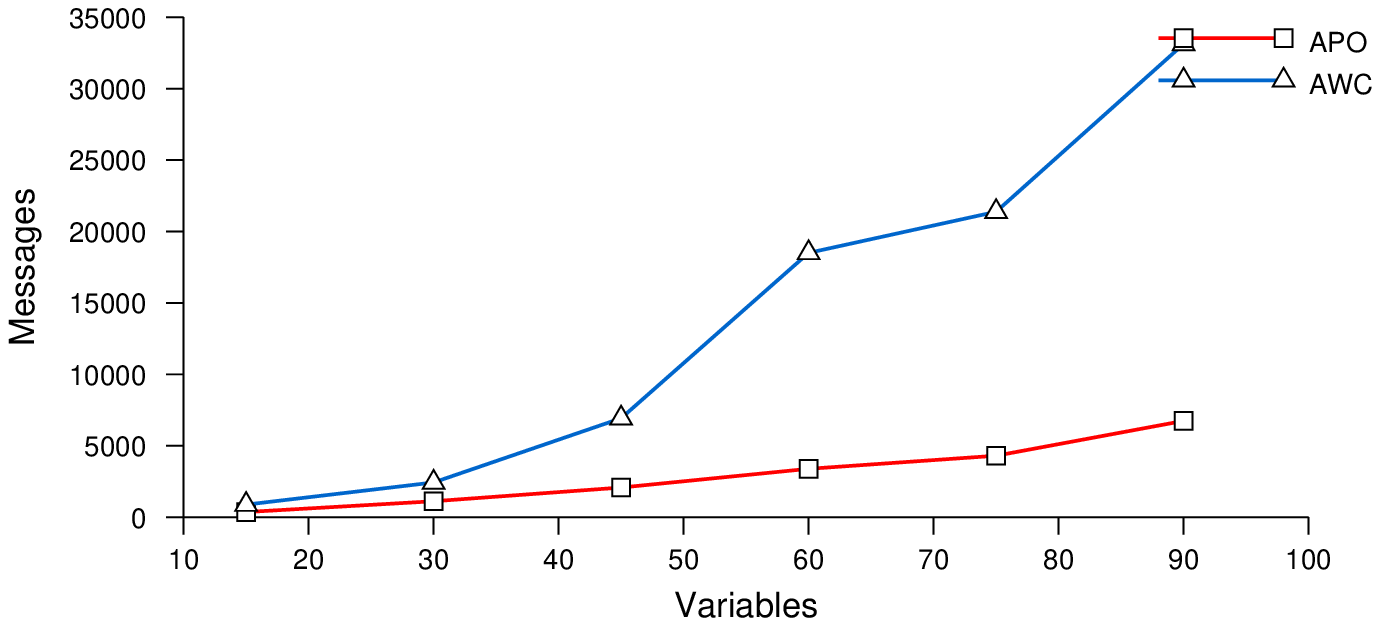}
  \caption{Comparison of the number of messages needed to solve satisfiable, low-density 3-coloring problems of various sizes by AWC and APO.}
   \label{satmess2.0}
\end{center}
\end{figure*}

\begin{figure*}
\begin{center}
   \epsfxsize=5.0in 
   \epsffile{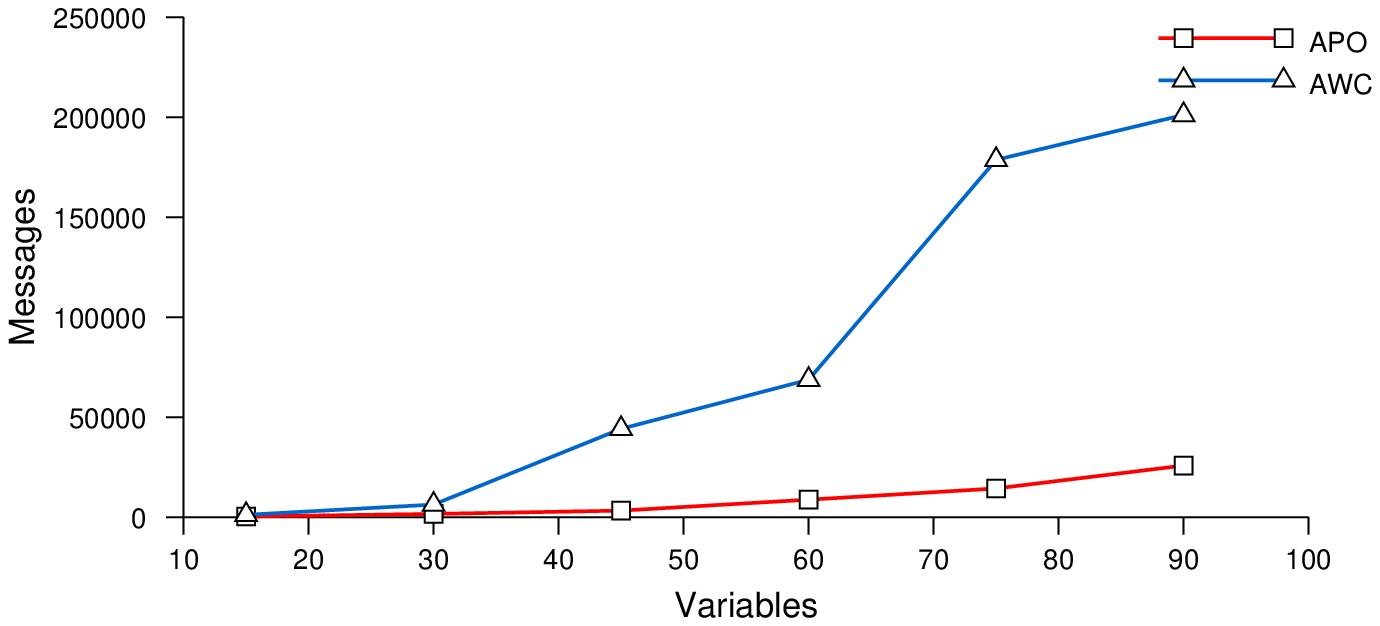}
  \caption{Comparison of the number of messages needed to solve satisfiable, medium-density 3-coloring problems of various sizes by AWC and APO.}
   \label{satmess2.3}
\end{center}
\end{figure*}

\begin{figure*}
\begin{center}
   \epsfxsize=5.0in 
   \epsffile{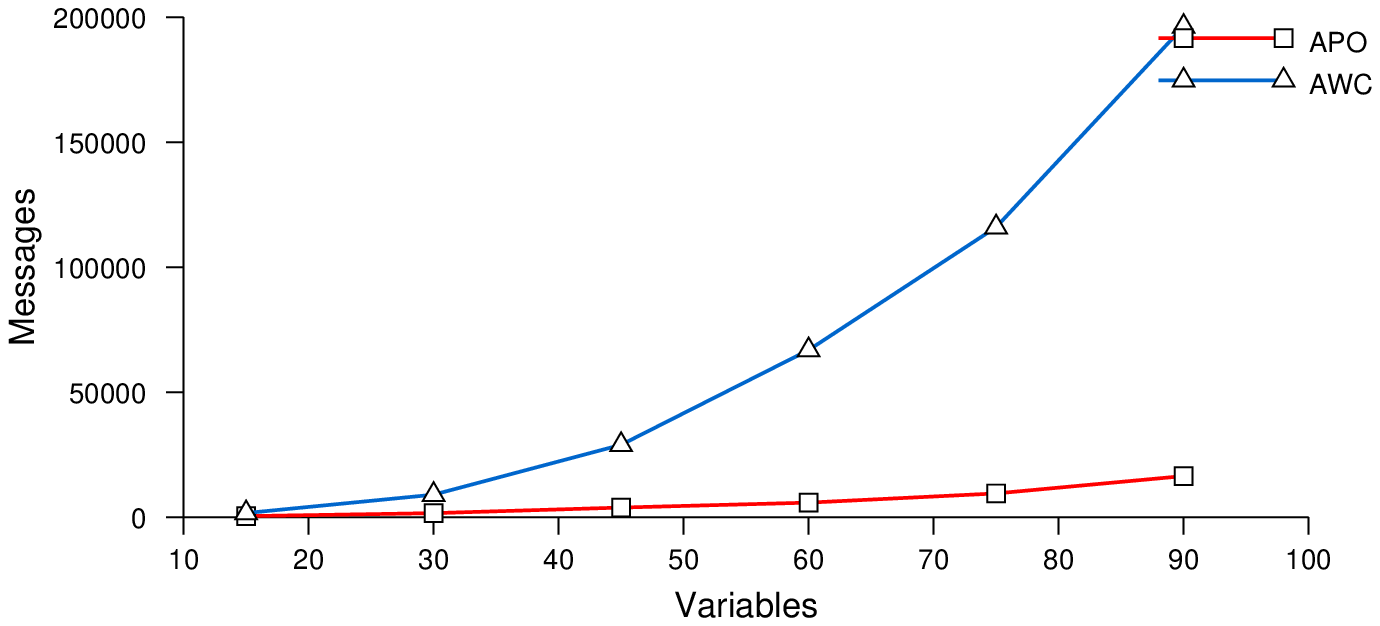}
  \caption{Comparison of the number of messages needed to solve satisfiable, high-density 3-coloring problems of various sizes by AWC and APO.}
   \label{satmess2.7}
\end{center}
\end{figure*}

For this comparison between AWC and APO, we randomly generated 10
graphs of size $n=15, 30, 45, 60, 75, 90$ and $m=2.0n,2.3n, 2.7n$ and
for each instance generated 10 initial variable assignments.
Therefore, for each combination of $n$ and $m$, we ran 100 trials
making a total of 1800 trials.  The results from this experiment can
be seen in figures \ref{sattime2.0} through \ref{satmess2.7} and table
\ref{table-satcyc}. We should mention that the results of the testing
on AWC obtained from these experiments agree with previous results
\cite{hirayama00learning} verifying the correctness of our
implementation.

At first glance, figure \ref{sattime2.0} appears to indicate that for
satisfiable low-density graph instances, AWC and APO perform almost
identically in terms of cycles to completion.  Looking at the
associated table (Table \ref{table-satcyc}), however, reveals that
overall, the pairwise T-test indicates that with 99\% confidence, APO
outperforms AWC on these graphs.

As the density, or average degree, of the graph increases, the
difference becomes more apparent. Figures \ref{sattime2.3} and
\ref{sattime2.7} show that APO begins to scale much more
efficiently than AWC.  This can be attributed to the ability of APO to
rapidly identify strong interdependencies between variables and to
derive solutions to them using a centralized search of the partial
subproblem.

\begin{table*}
\begin{center}

   \begin{tabular}{|c|c|r|r|r|r|r|}
      \hline
       & & \multicolumn{1}{c|}{APO}&  \multicolumn{1}{c|}{APO} &  \multicolumn{1}{c|}{AWC} & \multicolumn{1}{c|}{AWC} &\\ 
        $d$  & Nodes & \multicolumn{1}{c|}{Mean} &  \multicolumn{1}{c|}{StDev} & \multicolumn{1}{c|}{Mean} & \multicolumn{1}{c|}{StDev} & p($AWC \leq APO$)\\
      \hline
            & 15 & 361.50 & 179.57 & 882.19 & 967.21 & 0.00 \\ 
            & 30 & 1117.15 & 844.33 & 2431.71 & 3182.51 & 0.00 \\
        2.0 & 45 & 2078.72 & 1552.98 & 6926.86 & 7395.22 & 0.00 \\
            & 60 & 3387.13 & 2084.69 & 18504.17 & 22281.59 & 0.00 \\
            & 75 & 4304.22 & 2651.15 & 21219.01 & 22714.98 & 0.00 \\
            & 90 & 6742.14 & 4482.54 & 33125.57 & 39766.56 & 0.00 \\
            \cline{2-7}
	    & Overall &  &  &  &  & 0.01\\
      \hline
            & 15 & 379.15 & 188.69 & 1205.50 & 923.85 & 0.00 \\ 
            & 30 & 1640.08 & 931.22 & 6325.21 & 6914.79 & 0.00 \\
        2.3 & 45 & 3299.05 & 2155.45 & 44191.89 & 44693.33 & 0.00 \\
            & 60 & 8773.16 & 9613.84 & 70104.74 & 69050.66 & 0.00 \\
            & 75 & 14368.87 & 12066.32 & 178683.02 & 173493.21 & 0.00 \\
            & 90 & 25826.74 & 29172.66 & 201145.37 & 143236.26 & 0.00 \\
            \cline{2-7}
	    & Overall &  &  &  &  & 0.00\\
      \hline
            & 15 & 433.64 & 164.12 & 1667.71 & 1301.03 & 0.00 \\ 
            & 30 & 1623.89 & 787.59 & 9014.02 & 8104.34 & 0.00 \\
        2.7 & 45 & 3859.99 & 1921.51 & 28964.43 & 22900.89 & 0.00 \\
            & 60 & 5838.36 & 3140.53 & 66857.87 & 53221.05 & 0.00 \\
            & 75 & 9507.60 & 4486.04 & 116016.71 & 82857.63 & 0.00 \\
            & 90 & 16455.59 & 10679.92 & 196239.22 & 163722.90 & 0.00 \\
            \cline{2-7}
	    & Overall &  &  &  &  & 0.00\\
      \hline
   \end{tabular}
\end{center}
   \caption{Comparison of the number of messages needed to solve satisfiable 3-coloring problems of various sizes and densities by AWC and APO.}
   \label{table-satmess}
\end{table*}

\begin{table*}[tp]
\begin{center}

   \begin{tabular}{|c|c|r|r|r|r|r|}
      \hline
       & & \multicolumn{1}{c|}{APO}&  \multicolumn{1}{c|}{APO} &  \multicolumn{1}{c|}{AWC} & \multicolumn{1}{c|}{AWC} &\\ 
        $d$  & Nodes & \multicolumn{1}{c|}{Mean} &  \multicolumn{1}{c|}{StDev} & \multicolumn{1}{c|}{Mean} & \multicolumn{1}{c|}{StDev} & p($AWC \leq APO$)\\
      \hline
	    & 15 & 10560.76 & 4909.09 & 11590.38 & 13765.19 & 0.49\\
	    & 30 & 32314.97 & 23138.55 & 32409.93 & 45336.65 & 0.98\\
	2.0 & 45 & 59895.10 & 42740.82 & 95061.77 & 106391.51 & 0.00\\
	    & 60 & 97126.79 & 57428.70 & 259529.42 & 326087.70 & 0.00\\
	    & 75 & 123565.94 & 73493.72 & 294502.71 & 328696.59 & 0.00\\
	    & 90 & 192265.35 & 123384.05 & 466084.60 & 581963.72 & 0.00\\
            \cline{2-7}
	    & Overall &  &  &  &  & 0.00\\
      \hline
	    & 15 & 11370.13 & 4951.75 & 16260.19 & 13237.31 & 0.00\\
	    & 30 & 47539.20 & 25486.74 & 88946.59 & 101077.04 & 0.00\\
	2.3 & 45 & 95098.49 & 59312.25 & 644007.01 & 675192.26 & 0.00\\
	    & 60 & 247417.78 & 262844.89 & 1018059.11 & 1029273.23 & 0.00\\
	    & 75 & 401618.24 & 327990.65 & 2626178.31 & 2606377.80 & 0.00\\
	    & 90 & 712035.13 & 782835.83 & 2935211.45 & 2138087.17 & 0.00\\ 
            \cline{2-7}
	    & Overall &  &  &  &  & 0.00\\
      \hline
	    & 15 & 13415.51 & 4280.15 & 22393.61 & 18578.25 & 0.00\\
	    & 30 & 48542.24 & 21331.96 & 125072.76 & 116518.80 & 0.00\\
	2.7 & 45 & 112541.02 & 52729.10 & 405535.81 & 327349.47 & 0.00\\
	    & 60 & 170174.55 & 85705.12 & 945039.32 & 773937.60 & 0.00\\
	    & 75 & 272391.95 & 122177.07 & 1641250.63 & 1204185.34 & 0.00\\
	    & 90 & 465571.42 & 288265.82 & 2793725.78 & 2397839.47 & 0.00\\
            \cline{2-7}
	    & Overall &  &  &  &  & 0.00\\
      \hline
   \end{tabular}
\end{center}

   \caption{Comparison of the number of bytes transmitted by APO and AWC for satisfiable graph instances of various sizes and density.}
   \label{table-satbytes}
\end{table*}

Tables \ref{table-satlinks2.0} through \ref{table-satlinks2.7},
partially verify this statement.  As you can see, on average, less
than 50\% of the possible number of links ($n*(n-1)$) are used by APO
in solving problems (\% Links column).  In addition, the maximum
amount of centralization (\% Central column) occuring within any
single agent (i.e. the number of agents in its $agent\_view$) remains
fairly low.  The highest degree of centralization occurs for small,
high-density graphs.  Intuitively, this makes a lot of sense because
in these graphs, a single node is likely to have a high degree from
the very start.  Combine this fact with the dynamic priority ordering
and the result is large amounts of central problem solving.

The most profound differences in the algorithms can be seen in figures
\ref{satmess2.0}, \ref{satmess2.3}, and \ref{satmess2.7} and table
\ref{table-satmess}. APO uses at least an order of magnitude less
messages than AWC.  Table \ref{table-satbytes} shows that these
message savings lead to large savings in the number of bytes being
transmitted as well.  Even though APO uses about twice as many bytes
per message as AWC (the messages were not optimized at all), the total
amount of information be passed around is significantly less in almost
every case.

Again, looking at the linking structure that AWC produces gives some
insights into why it uses some many more messages than APO.  Because
agents communicate with all of the agents they are linked to whenever
their value changes, and a large number of changes can occur in single
cycle, AWC can have a tremendous amount of thrashing behavior.  APO,
on the other hand, avoids this problem because the process of
mediating implicitly creates regions of stability in the problem
landscape while the mediator decides on a solution.  In addition,
because APO uses partial centralization to solve problems, it avoids
having to use a large number of messages to discover implied
constraints through trial and error.

As we will see in the next two experiments, the high degree of
centralization caused by its unfocused linking degrades AWC's
performance even more when solving randomly generated, possibly
unsatisfiable, graph instances.

\begin{figure}
\begin{center}
   \epsfxsize=4.5in \epsffile{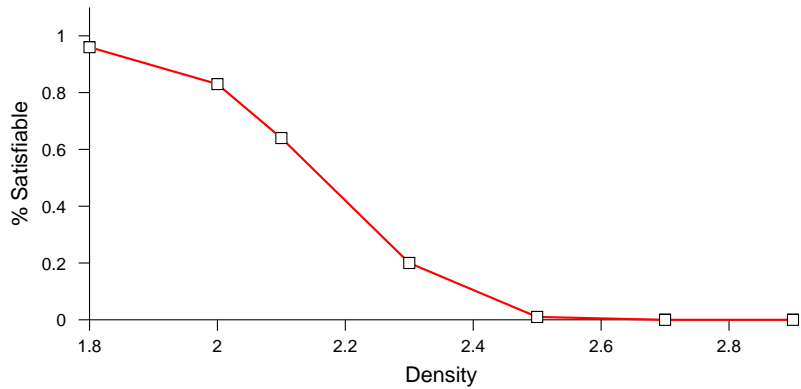}
\end{center}
   \caption{Phase transition curve for the 60 node randomly generated
   graphs used in testing.}  \label{any-phase}
\end{figure}

\begin{figure*}
\begin{center}
   \epsfxsize=5.0in 
   \epsffile{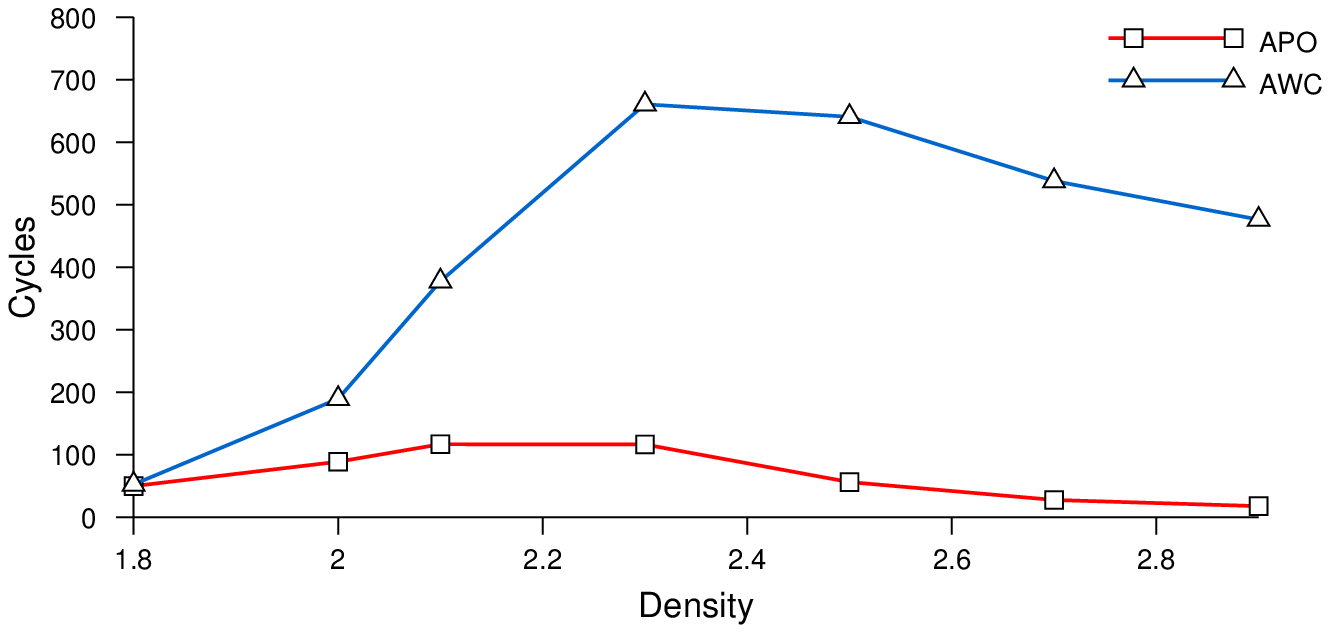}
  \caption{Number of cycles needed to solve completely random 60 variable problems of various density using AWC and APO.}
   \label{atime}
\end{center}
\end{figure*}

\begin{table}
\begin{center}
    \begin{tabular}{|r|r|r|r|r|r|r|r|}
      \hline
      & \multicolumn{1}{c|}{APO}&  \multicolumn{1}{c|}{APO} &  \multicolumn{1}{c|}{\% APO} & \multicolumn{1}{c|}{AWC} & \multicolumn{1}{c|}{AWC} &  \multicolumn{1}{c|}{\% AWC} & \\ 
       Density & \multicolumn{1}{c|}{Mean} &  \multicolumn{1}{c|}{StDev} & \multicolumn{1}{c|}{Solved} & \multicolumn{1}{c|}{Mean} &  \multicolumn{1}{c|}{StDev} & \multicolumn{1}{c|}{Solved} & $p(AWC \leq APO)$\\
      \hline
      1.8 & 49.88 & 41.98 & 100 & 52.51 & 77.06 & 100 & 0.62 \\ 
      2.0 & 88.77 & 83.79 & 100 & 189.42 & 237.17 & 96 & 0.00 \\
      2.1 & 116.79 & 107.21 & 100 & 377.54 & 364.55 & 80 & 0.00 \\
      2.3 & 116.41 & 264.65 & 100 & 660.65 & 362.80 & 55 & 0.00 \\
      2.5 & 56.21 & 45.12 & 100 & 640.66 & 335.65 & 65 & 0.00 \\
      2.7 & 27.62 & 25.66 & 100 & 537.99 & 324.87 & 83 & 0.00\\
      2.9 & 17.74 & 13.69 & 100 & 476.20 & 271.63 & 92 & 0.00 \\
      \hline
      Overall & & & & & & &0.00 \\
      \hline
   \end{tabular}
  \caption{Number of cycles needed to solve completely random 60 variable problems of various density using AWC and APO.}
   \label{tabatime}
\end{center}
\end{table}

\subsubsection{Random Graphs}
In the second set of experiments, we generated completely random 60
node graphs with average degrees from $d = 1.8$ to $2.9$.  This series
was conducted to test the completeness of the algorithms, verify the
correctness of their implementations, and to study the effects of the
phase transition on their performance.  For each value of $d$, we
generated 200 random graphs each with a single set of initial values.
Graphs were generate by randomly choosing two nodes and connecting
them.  If the edge already existed, another pair was chosen.  The
phase transition curve for these instances can be seen in figure
\ref{any-phase}.

\begin{figure*}
\begin{center}
   \epsfxsize=5.0in 
   \epsffile{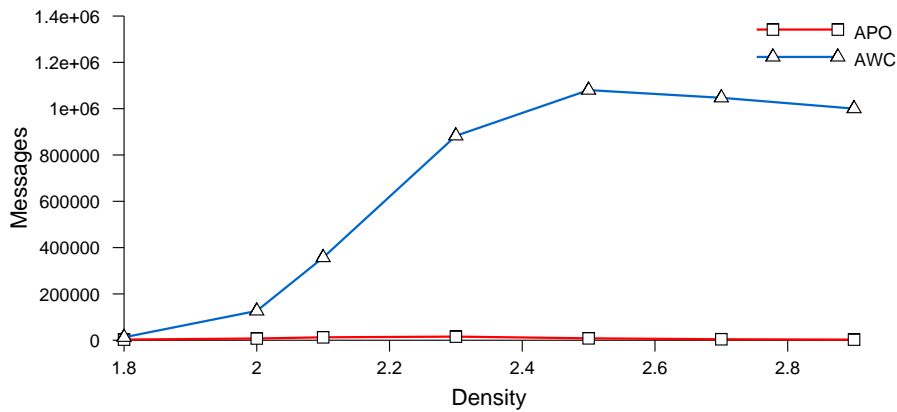}
  \caption{Number of messages needed to solve completely random 60 variable problems of various density using AWC and APO.}
   \label{amess}
\end{center}
\end{figure*}

In total, 1400 graphs were generated and tested.  Due to time
constraints, we stopped the execution of AWC once 1000 cycles had
completed (APO never reached 1000).  The results of these experiments
are shown in figures \ref{atime} and \ref{amess} and tables
\ref{tabatime} and \ref{tabamess}.

\begin{table}
\begin{center}
   \begin{tabular}{|r|r|r|r|r|r|}
      \hline
      & \multicolumn{1}{c|}{APO}&  \multicolumn{1}{c|}{APO} & \multicolumn{1}{c|}{AWC} & \multicolumn{1}{c|}{AWC} & \\ 
       Density & \multicolumn{1}{c|}{Mean} &  \multicolumn{1}{c|}{StDev} & \multicolumn{1}{c|}{Mean} &  \multicolumn{1}{c|}{StDev} & $p(AWC \leq APO)$\\ 
      \hline
      1.8 & 2822.61 & 3040.39 & 12741.58 & 47165.70 & 0.00 \\ 
      2.0 & 7508.33 & 9577.86 & 126658.29 & 269976.18 & 0.00 \\
      2.1 & 12642.68 & 16193.56 & 356993.39 & 444899.21 & 0.00 \\
      2.3 & 15614.37 & 15761.90 & 882813.45 & 566715.73 & 0.00 \\
      2.5 & 8219.74 & 7415.76 & 1080277.25 & 661909.63 & 0.00 \\
      2.7 & 4196.58 & 4201.80 & 1047001.18 & 738367.27 & 0.00\\
      2.9 & 2736.20 & 2286.39 & 1000217.83 & 699199.90 & 0.00 \\
      \hline
      Overall & & & & & 0.00 \\
      \hline
   \end{tabular}
  \caption{Number of messages needed to solve completely random 60 variable problems of various density using AWC and APO.}
   \label{tabamess}
\end{center}
\end{table}

\begin{table*}[t]
   \begin{center}
   \begin{tabular}{|c|c|r|r|r|r|}
      \hline
       & & \multicolumn{1}{c|}{\% Links}&  \multicolumn{1}{c|}{\% Links} &  \multicolumn{1}{c|}{\% Central} & \multicolumn{1}{c|}{\% Central} \\ 
       & Density & \multicolumn{1}{c|}{Mean} &  \multicolumn{1}{c|}{StDev} & \multicolumn{1}{c|}{Mean} & \multicolumn{1}{c|}{StDev} \\
      \hline
      & 1.8 & 8.09 & 1.50 & 26.19 & 6.94 \\ 
      & 2.0 & 10.93 & 3.28 & 36.92 & 13.77 \\ 
      & 2.1 & 13.31 & 4.49 & 46.68 & 15.86 \\ 
      APO & 2.3 & 15.56 & 4.94 & 55.91 & 18.18 \\ 
      & 2.5 & 14.37 & 3.74 & 53.86 & 18.25 \\ 
      & 2.7 & 13.19 & 3.32 & 47.33 & 19.59 \\ 
      & 2.9 & 13.12 & 2.70 & 45.26 & 17.55 \\
      \hline
      & 1.8 & 16.28 & 6.58 & 41.08 & 12.42 \\  
      & 2.0 & 34.34 & 13.44 & 65.00 & 14.41 \\ 
      & 2.1 & 46.52 & 13.95 & 75.24 & 10.66 \\ 
      AWC & 2.3 & 64.13 & 11.43 & 86.06 & 5.61 \\ 
      & 2.5 & 70.06 & 10.51 & 89.46 & 4.55 \\ 
      & 2.7 & 72.95 & 10.65 & 91.71 & 3.80 \\ 
      & 2.9 & 75.19 & 9.62 & 92.78 & 5.63 \\
      \hline
   \end{tabular}
\end{center}
   \caption{Link statistics for 60 node random problems.}
   \label{table-any}
\end{table*}

On these graphs, APO significantly outperforms AWC on all but the
simplest problems (see figure \ref{atime}).  These results can most
directly be attributed to AWC's poor performance on unsatisfiable
problem instances \cite{selman03dsn}.  In
fact, in the region of the phase transition, AWC was unable to
complete 45\% of the graphs within the 1000 cycles.

In addition, to solve these problems, AWC uses at least an order of
magnitude more messages that APO.  These results can be seen in figure
\ref{amess}.  By looking at table \ref{table-any}, it is easy to see
why this occurs.  AWC has a very high degree of linking and
centralization.  In fact, on $d=2.9$ graphs, AWC reaches an average of
93\% centralization and 75\% of complete inter-agent linking.

To contrast this, APO has very loose linking throughout the entire
phase transition and centralizes on average around 50\% of the entire
problem.  These results are very encouraging and reinforce the idea
that partial overlays and extending along critical paths yields
improvements in the convergence on solutions.

\subsubsection{Runtime Tests}
\label{APO:rt-test}
\begin{figure*}
   \epsfxsize=5.8in 
   \epsffile{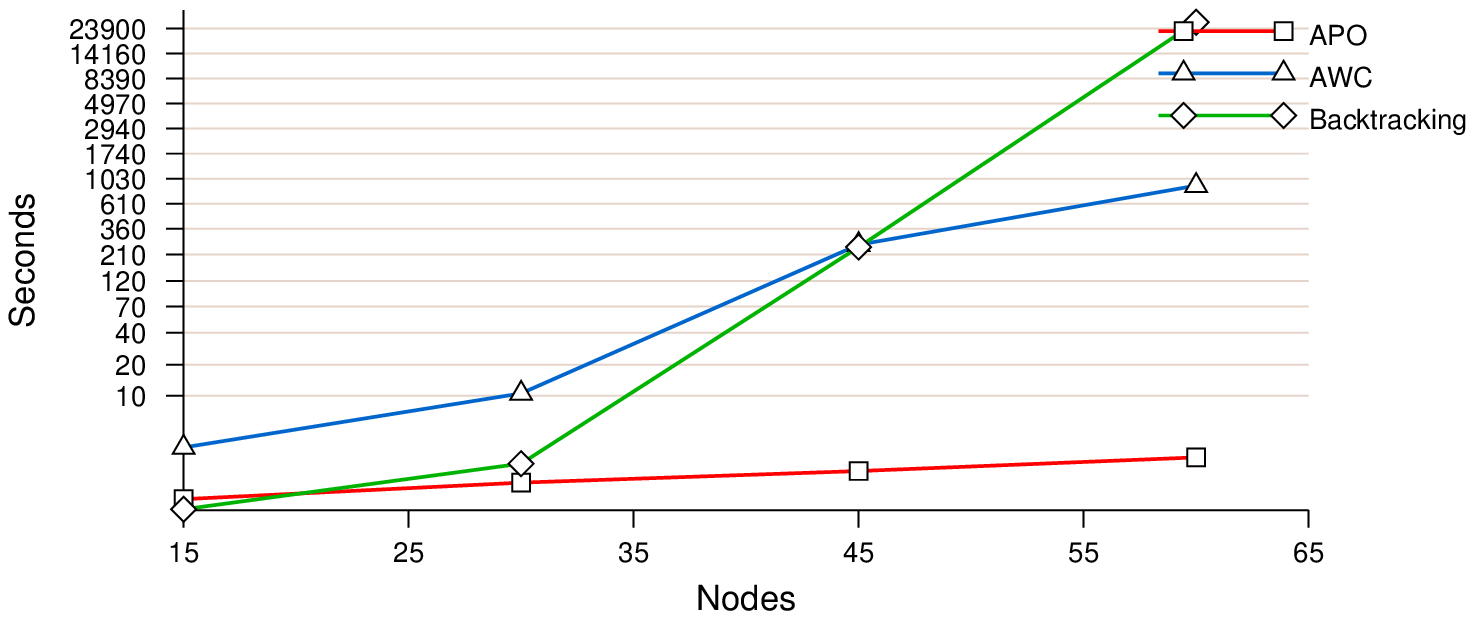}
    \caption{Comparison of the number of seconds needed to solve random, low-density 3-coloring problems of various sizes by AWC, APO, and centralized Backtracking.}
   \label{anyrt2.0}
\end{figure*}

\begin{figure*}[tp]
   \epsfxsize=5.8in 
   \epsffile{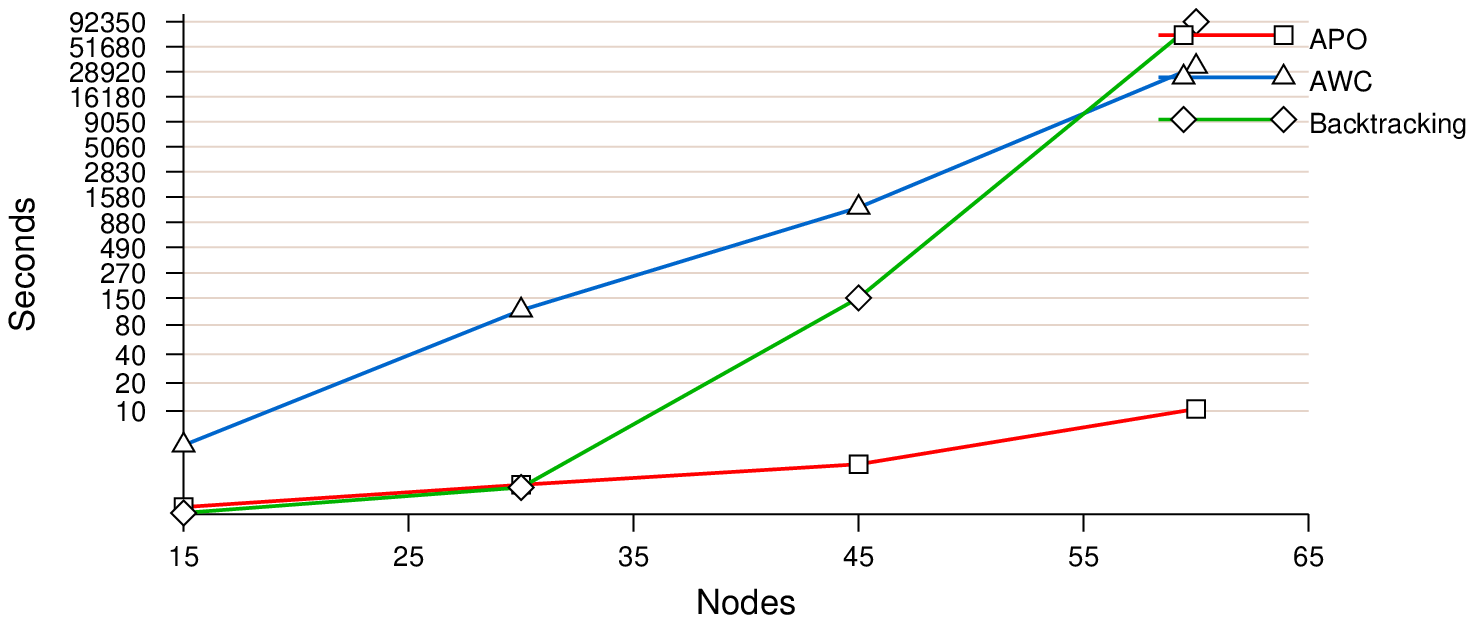}
  \caption{Comparison of the number of seconds needed to solve random, medium-density 3-coloring problems of various sizes by AWC, APO, and centralized Backtracking.}
   \label{anyrt2.3}
\end{figure*}

\begin{figure*}[tp]
   \epsfxsize=5.8in 
   \epsffile{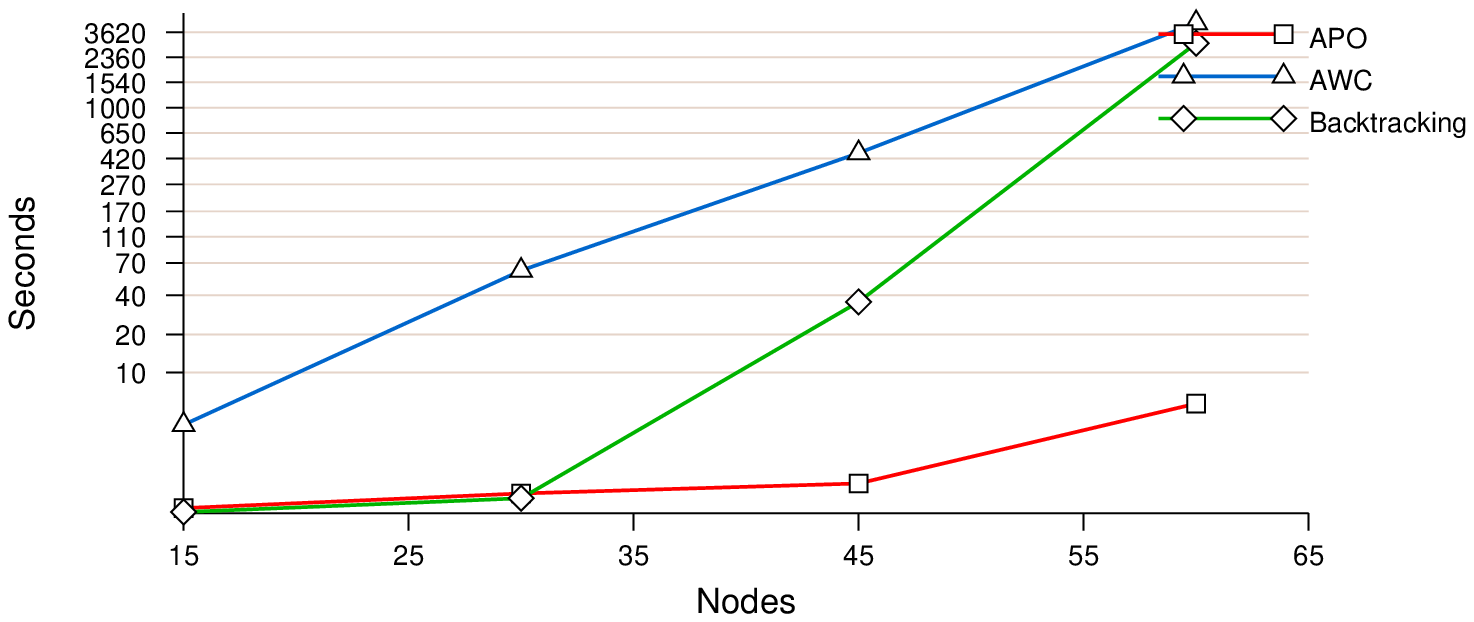}
  \caption{Comparison of the number of seconds needed to solve random, high-density 3-coloring problems of various sizes by AWC, APO, and centralized Backtracking.}
   \label{anyrt2.7}
\end{figure*}

\begin{table*}
\begin{center}

   \begin{tabular}{|c|c|r|r|r|r|r|r|}
      \hline
      & & \multicolumn{1}{c|}{APO}& \multicolumn{1}{c|}{APO} & \multicolumn{1}{c|}{AWC} & \multicolumn{1}{c|}{AWC} & \multicolumn{1}{c|}{BT} & \multicolumn{1}{c|}{BT}\\
      $d$ & Nodes & \multicolumn{1}{c|}{Mean}& \multicolumn{1}{c|}{StDev} & \multicolumn{1}{c|}{Mean} & \multicolumn{1}{c|}{StDev} & \multicolumn{1}{c|}{Mean} & \multicolumn{1}{c|}{StDev} \\ 

      \hline
	    & 15 & 0.26 & 0.21 & 2.72 & 4.40 & 0.02 & 0.02\\
        2.0 & 30 & 0.78 & 0.66 & 10.52 & 18.99 & 1.65 & 4.14\\
            & 45 & 1.27 & 1.18 & 257.88 & 1252.25 & 245.14 & 587.31\\
	    & 60 & 2.02 & 1.74 & 890.12 & 4288.43 & 27183.64 & 56053.26\\
      \hline
	    & 15 & 0.18 & 0.23 & 3.96 & 3.23 & 0.03 & 0.01\\
	2.3 & 30 & 0.98 & 0.88 & 112.95 & 144.24 & 0.86 & 1.07\\
	    & 45 & 2.19 & 2.26 & 1236.27 & 1777.51 & 150.73 & 241.02\\
	    & 60 & 10.51 & 12.97 & 32616.24 & 62111.52 & 92173.71 & 222327.59\\
      \hline
	    & 15 & 0.09 & 0.06 & 3.51 & 2.90 & 0.02 & 0.01\\
	2.7 & 30 & 0.40 & 0.39 & 61.43 & 59.77 & 0.29 & 0.36\\
	    & 45 & 0.66 & 0.63 & 460.56 & 690.98 & 35.58 & 57.75\\
	    & 60 & 5.47 & 5.44 & 4239.16 & 4114.30 & 2997.04 & 4379.97\\
      \hline
   \end{tabular}
\end{center}
   \caption{Comparison of the number of seconds needed to solve random 3-coloring problems of various sizes and densities using AWC, APO, and centralized Backtracking.}
   \label{table-anysrt}
\end{table*}

In the third set of experiments, we directly compared the serial
runtime performance of AWC against APO.  Serial runtime is measured
using the following formula:

\[serial time = \sum_{i=0} ^{cycles}\sum_{a \in agents} time(a,i)\]

Which is just the total accumulated runtime needed to solve the problem
when only one agent is allowed to process at a time.

For these experiments, we again generated random graphs, this time
varying the size and the density of the graph.  We generated 25 graphs
for the values of $n=15,30,45,60$ and the densities of $d = 2.0, 2.3,
2.7$, for a total of 300 test cases.  To show that the performance
difference in APO and AWC was not caused by the speed of the central
solver, we ran a centralized backtracking algorithm on the same graph
instances.  Although, APO uses the branch and bound algorithm, the
backtracking algorithm used in this test provides a best case lower
bound on the runtime of APO's internal solver.

Each of the programs used in this test was run on an identical 2.4GHz
Pentium 4 with 768 Mbytes of RAM.  These machines where entirely
dedicated to the tests so there was a minimal amount of interference
by competing processes.  In addition, no computational cost was
assigned to message passing because the simulator passes messages
between cycles.  The algorithms were, however, penalized for the
amount of time they took to process messages.  Although we realize
that the specific implementation of an algorithm can greatly effect
its runtime performance, every possible effort was made to optimize
the AWC implementation used for these experiments in an effort to be
fair.

The results of this test series can be seen in figures \ref{anyrt2.0},
\ref{anyrt2.3}, and \ref{anyrt2.7}.  You should note that the scale
used for these graphs is logarithmic.  From looking at these results,
two things should become apparent.  Obviously, the first is that APO
outperforms AWC in every case.  Second, APO actually outperforms its
own centralized solver on graphs larger than 45 nodes.  This indicates
two things.  First, the solver that is currently in APO is very poor
and second that APO's runtime performance is not a direct result of
the speed of centralized solver that it is using.  In fact, these
tests show that the improved performance of APO over AWC is caused by
APO's ability to take advantage of the problem's structure.

If we were to replace the centralized solver used in these tests with
a state-of-the-art solver, we would expect two things.  The first is
that we would expect the serial runtime of the APO algorithm to
decrease simply from the speedup caused by the centralized solver.
The second, and more importantly, is that the centralized solver would
always outperform APO.  This is because current CSP solvers take
advantage of problem structure unlike the solver used in these tests.
We are in no way making a claim that APO improves any centralized
solver.  We are simply stating that APO outperforms AWC for reasons
other than the speed of its current internal solver.

\begin{figure}
\begin{center}
   \epsfxsize=4.5in \epsffile{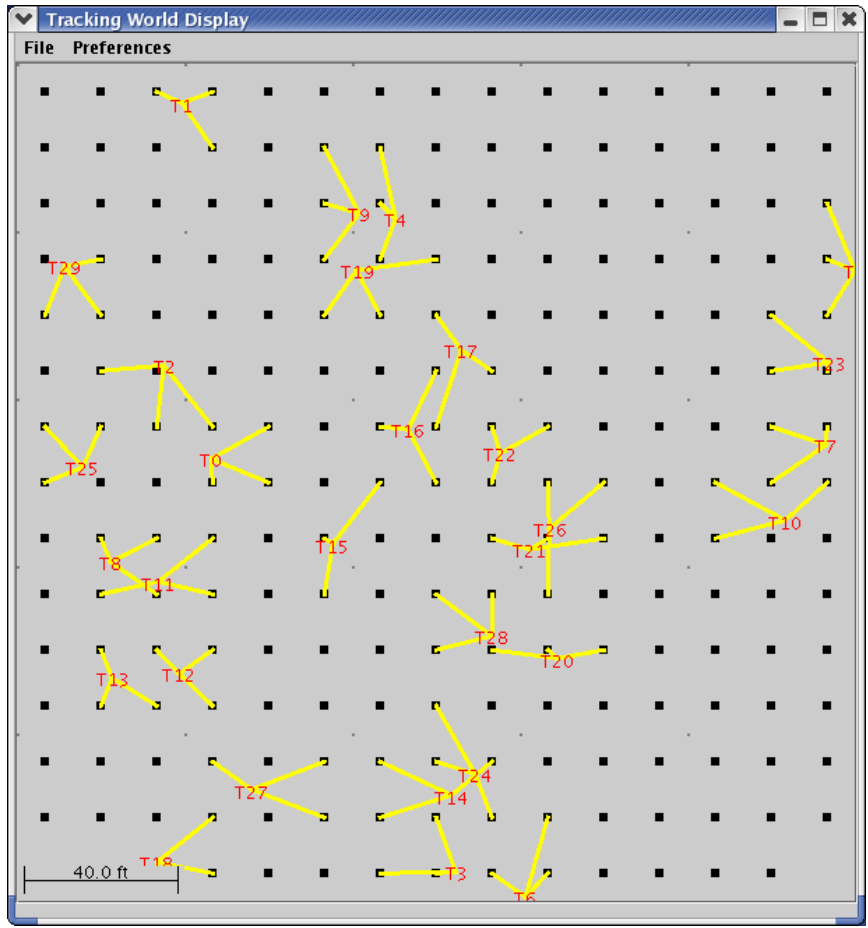}
\end{center}
   \caption{An example of a tracking problem.  There are 30 targets
   (labeled with their name) and 224 sensors (black dots).  Lines
   connecting sensors and targets indicate that the sensor is assigned
   to tracking the target.}  \label{trackdisplay}
\end{figure}

\subsection{Tracking Domain}
\label{tracking}
To test APO's adaptability to various centralized solvers, we created
an implementation of the complete-compatibility version of the
SensorDCSP formulation 
\cite{bejar01distributed,krishnamachari02distributed,selman03dsn}.  In
this domain, there are a number of sensors and a number of targets
randomly placed within an environment.  Because of range restrictions,
only sensors that are within some distance $dist$ can see a target.
The goal is to find an assignment of sensors to targets such that each
target has three sensors tracking it.

Following directly from the definition for a CSP, a SensorDCSP problem
consists of the following:
\begin{itemize}
\item a set of {\it n} targets $T = \{T_1,\ldots,T_n\}$.
\item a set of possible sensors that can ``see'' each of the targets $D = \{D_1,\ldots,D_n\}$.
\item a set of constraints $R = \{R_1,\ldots,R_m\}$ where each $R_i(a_i,a_j)$ is predicate which implements the ``not intersects'' relationship.  This predicate returns true iff the sensors assigned to $T_i$ does not have any elements in common with the sensors assigned to $T_j$.
\end{itemize}

The problem is to find an assignment $A=\{a_1,\ldots,a_n\}$ such that
each of the constraints in $R$ is satisfied and each $a_i$ is a set of
$|D_i| \choose c$ sensors from $D_i$ where $c =min(|D_i|,3)$.  This
indicates that each target requires 3 sensors, if enough are
available, or all of the sensors, if there are less than 3.

Since, in this implementation, each of the sensors is compatible with
one another, the overall complexity of the problem is polynomial, using a reduction to
feasible flow in a bipartite graph\cite{krishnamachari02thesis}.  Because of this, the centralized
solver used by the APO agents was changed to a modified version of the
Ford-Fulkerson maximum flow algorithm
\cite{ford62,clr99}, which has been proven to run in polynomial time.

An example of the tracking problem can be seen in figure
\ref{trackdisplay}.  In this example, there are 224 sensors (black
dots) placed in an ordered pattern in the environment.  There are 30
targets (labeled with their names) which are randomly placed at
startup.  The lines connecting sensors to targets indicate that the
sensor is assigned to the target.  Note that this instance of the
problem is satisfiable.
 
\subsubsection{Modifying APO for the Tracking Domain}
Because the tracking domain is so closely related to the general CSP
formulation, very few changes were made to either AWC or APO for these
tests.  We did, however, decide to test the adaptability of APO to a
new centralized problem solver.  To do this, we changed the
centralized problem solver to the Ford Fulkerson max-flow algorithm in
figure \ref {max_flow}.  Ford-Fulkerson works by repeatedly finding
paths with remaining capacity through the residual flow network and
augmenting the flows along those paths.  The algorithm terminates when
no additional paths can be found.  A detailed explanation of the
algorithm as well as a proof of its optimality can be found in
\cite{clr99}.

\begin{figure*}
\centerline{
\hbox{
\progstyle{
{\bf Ford-Fulkerson} $(G, s, t)$ \nnl
\> {\bf for each} edge $(u,v) \in E[G]$ {\bf do} \nnl
\>\> $f[u,v] \leftarrow 0$; \nnl
\>\> $f[v,u] \leftarrow 0$; \nnl
\> {\bf end do}; \nnl\nnl
\> {\bf while} there exists a path $p$ from $s$ to $t$ \nnl
\>\>\ \ \ in the residual network $G_f$ {\bf do} \nnl
\>\> $c_f(p) \leftarrow$ min$\{c_f(u,v):(u,v) \in p\}$; \nnl
\>\> {\bf for each} edge $(u,v) \in p$ {\bf do} \nnl
\>\>\> $f[u,v] \leftarrow f[u,v]+ c_f(p)$; \nnl
\>\>\> $f[v,u] \leftarrow -f[u,v]$; \nnl
\>\> {\bf end do}; \nnl
\> {\bf end do}; \nnl
{\bf end Ford-Fulkerson};
}
}
}
\caption{The Ford-Fulkerson maximum flow algorithm.}
\label{max_flow}
\end{figure*}

\begin{figure}
\begin{center}
   \epsfxsize=1.6in \epsffile{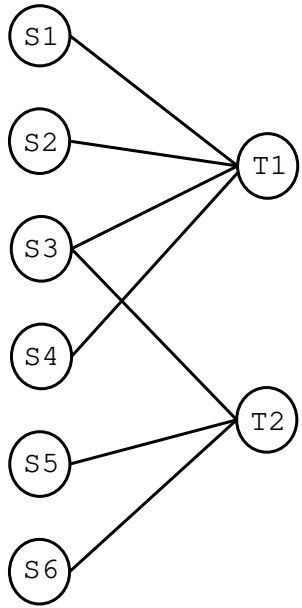}
\end{center}
   \caption{A simple sensor to target allocation problem.}
   \label{maxflow1}
\end{figure}

\begin{figure}
\begin{center}
   \epsfxsize=3.5in \epsffile{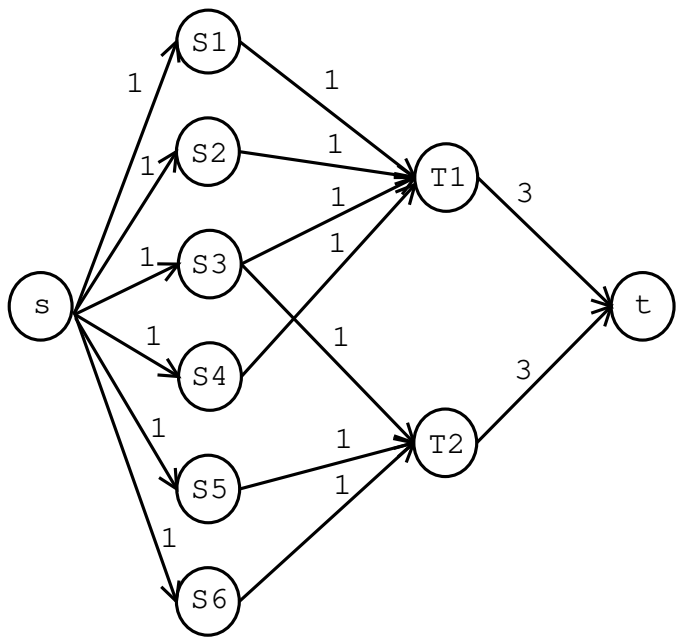}
\end{center}
   \caption{The flow network for the simple target allocation problem
   in figure \ref{maxflow1}.}  
\label{maxflow2}
\end{figure}

Like mapping bipartite graphs into max-flow, the SensorDCSP problem is
also easily mapped into max-flow.  In figures \ref{maxflow1} and
\ref{maxflow2} you can see the mapping of a simple sensor allocation
problem into a max-flow problem.  Notice that the capacity of the flow
between the sensors and targets is 1.  This ensures that a sensor
cannot be used by more than a single target.  Also, notice that the
capacity of the targets to $t$ is 3.  In fact, this value is to
min$(|D_i|,3)$.

To use this algorithm within APO, the mediator simply translates the
problem into a network flow graph $G$ using the following rules
whenever it runs the choose\_solution procedure in figure
\ref{apo_choose}:

\begin{enumerate}
\item Add the nodes $s$ and $t$ to $G$.
\item For each $T_i \in T$ add a node $T_i$ and an edge $(T_i, t)$ with capacity min$(|D_i|,3)$ to $G$.
\item For each unique sensor $S_i$ in the domains of $T_i \in T$, add a node $S_i$, an edge $(s, S_i)$ with capacity 1, and an edge $(S_i, T_i)$ with capacity 1 to $G$.
\end{enumerate}

It then executes the Ford-Fulkerson algorithm.  Once the algorithm
finishes, the mediator checks the residual capacity of the edges
between the targets and $t$.  If any of these edges has residual flow,
then the problem is unsatisfiable.  Otherwise, the assignment can be
derived by finding all of the $(S_i, T_i)$ edges that have a flow of
1.
 
One of the nicest characteristics of the Ford-Fulkerson algorithm is
that it works regardless of the order that the paths in the residual
network are chosen.  In our implementation, we used a breadth-first
search which, in addition to identifying paths in the residual
network, minimized the cost of the path.  Cost in this sense refers to
the amount of external conflict that is created by having a sensor
assigned to a target.  This modification maintains the min-conflict
heuristic which is an integral part of extending the mediators local
view.

\subsubsection{Results}
\begin{figure}
\begin{center}
   \epsfxsize=4.5in \epsffile{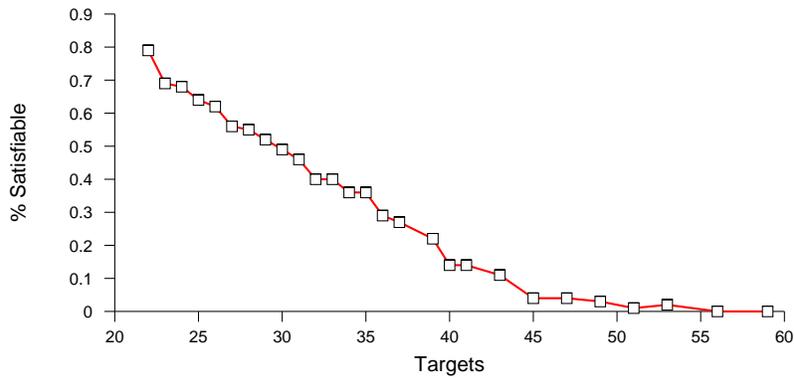}
\end{center}
   \caption{Phase transition curve for the 224 sensor environment used
   for testing.}  
\label{tracksat}
\end{figure}

To test APO and AWC in this domain, we ran a test series which used a
$200ft \times 200 ft$ environment with 224 sensors placed in an
ordered grid-based pattern.  We chose to place the sensors in an
ordered fashion to reduce the variance obtained within the results.
We ran a test series which varied the sensor to target ratio from 10:1
to 3.8:1 (22 to 59 targets) in increments of 0.2 which is across the
spectrum from mostly satisfiable to mostly unsatisfiable instances
(see figure \ref{tracksat}).  We then conducted 250 trial runs with a random
target placement for each of these values to get a good statistical
sampling.

In total, 6750 test cases were used.  For comparison, we measured the
number of messages and cycles that were taken by the algorithms to
find a solution.  The random seeds used to place the targets were
saved, so APO and AWC were both tested using identical problem
instances.  The correctness of the algorithms was verified by
cross-checking the solutions (satisfiable/unsatisfiable) obtained
during these tests, which matched identically.

\begin{figure*}[tp]
\begin{center}
   \epsfxsize=4.0in 
   \epsffile{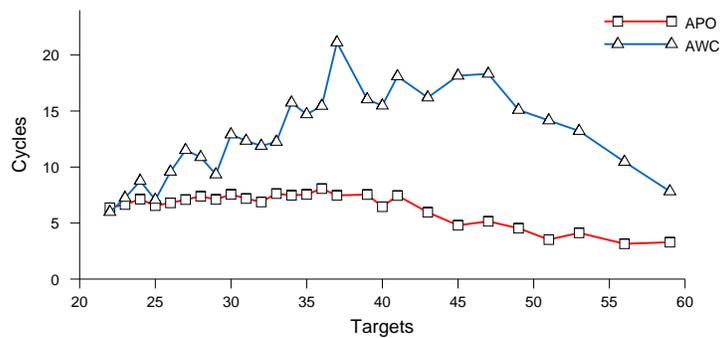}
  \caption{Number of cycles needed to solve random target configurations in a field of 224 sensors using AWC and APO.}
   \label{tracktime}
\end{center}
\end{figure*}

\begin{figure*}[tp]
\begin{center}
   \epsfxsize=4.02in 
   \epsffile{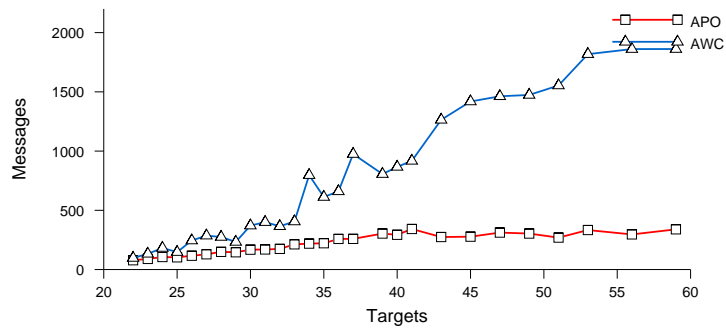}
   \caption{Number of messages needed to solve random target configurations targets in a field of 224 sensors using AWC and APO.}
   \label{trackmess}
\end{center}
\end{figure*}

As can be seen in figure \ref{tracktime} and \ref{trackmess} and tables \ref{tabtracktime} and \ref{tabtrackmess}, APO
outperforms AWC on all but the simplest cases.  Part of the reason for
this is the minimum 3 cycles it takes APO to finish a mediation
session.  In problems that have very sparsely connected
interdependencies, this cost tends to dominate.  All-in-all, as the
T-tests indicate, APO is significantly better than AWC in terms of
both cycles to completion and number of messages used for problems in
this domain.

\begin{table}
\begin{center}
   \begin{tabular}{|r|r|r|r|r|r|}
      \hline
  	& \multicolumn{1}{c|}{APO}& \multicolumn{1}{c|}{APO} & \multicolumn{1}{c|}{AWC} & \multicolumn{1}{c|}{AWC} & \\
      Targets & \multicolumn{1}{c|}{Mean}& \multicolumn{1}{c|}{StDev} & \multicolumn{1}{c|}{Mean} & \multicolumn{1}{c|}{StDev} &  $p(AWC \leq APO)$ \\ 
      \hline
      22 & 6.36 & 2.33 & 5.88 & 6.61 & 0.29 \\ 
      23 & 6.65 & 3.39 & 7.32 & 10.55 & 0.36 \\
      24 & 7.12 & 4.72 & 8.83 & 19.96 & 0.18 \\
      25 & 6.55 & 3.24 & 7.15 & 10.56 & 0.39 \\
      26 & 6.80 & 4.28 & 9.65 & 15.78 & 0.01 \\
      27 & 7.09 & 5.02 & 10.85 & 19.89 & 0.00 \\
      28 & 7.38 & 5.88 & 11.05 & 15.89 & 0.00 \\ 
      29 & 7.10 & 4.89 & 9.24 & 13.76 & 0.02 \\
      30 & 7.55 & 5.99 & 13.15 & 25.58 & 0.00 \\
      31 & 7.18 & 6.11 & 12.42 & 22.22 & 0.00 \\
      32 & 6.88 & 6.31 & 11.73 & 18.21 & 0.00 \\
      33 & 7.62 & 7.59 & 12.32 & 24.04 & 0.00 \\      
      34 & 7.47 & 7.81 & 15.88 & 28.82 & 0.00 \\
      35 & 7.56 & 7.49 & 14.74 & 29.01 & 0.00 \\ 
      36 & 8.08 & 9.89 & 15.70 & 25.46 & 0.00 \\
      37 & 7.48 & 8.38 & 20.70 & 39.63 & 0.00 \\
      39 & 7.55 & 10.87 & 16.12 & 26.43 & 0.00 \\
      40 & 6.45 & 10.54 & 15.74 & 21.66 & 0.00 \\
      41 & 7.45 & 13.11 & 17.56 & 30.98 & 0.00 \\
      43 & 5.96 & 7.79 & 16.10 & 22.91 & 0.00 \\
      45 & 4.80 & 7.25 & 17.61 & 28.50 & 0.00 \\
      47 & 5.15 & 8.31 & 18.52 & 30.27 & 0.00 \\      
      49 & 4.53 & 5.90 & 15.33 & 25.28 & 0.00 \\
      51 & 3.52 & 2.00 & 14.34 & 22.60 & 0.00 \\ 
      53 & 4.12 & 5.82 & 13.13 & 22.55 & 0.00 \\
      56 & 3.14 & 0.59 & 10.45 & 20.81 & 0.00 \\
      59 & 3.28 & 2.26 & 7.46 & 10.47 & 0.00 \\
      \hline
      Overall & & & & & 0.00 \\
      \hline
   \end{tabular}
  \caption{Number of cycles needed to solve random target configurations in a field of 224 sensors using AWC and APO.}
   \label{tabtracktime}
\end{center}
\end{table}

\begin{table}
\begin{center}
   \begin{tabular}{|r|r|r|r|r|r|}
      \hline
  	& \multicolumn{1}{c|}{APO}& \multicolumn{1}{c|}{APO} & \multicolumn{1}{c|}{AWC} & \multicolumn{1}{c|}{AWC} & \\
      Targets & \multicolumn{1}{c|}{Mean}& \multicolumn{1}{c|}{StDev} & \multicolumn{1}{c|}{Mean} & \multicolumn{1}{c|}{StDev} &  $p(AWC \leq APO)$ \\ 
      \hline
      22 & 78.28 & 32.58 & 95.68 & 133.53 & 0.04 \\ 
      23 & 89.52 & 54.01 & 133.12 & 237.74 & 0.00 \\
      24 & 105.53 & 90.36 & 184.19 & 616.36 & 0.04 \\
      25 & 102.92 & 57.18 & 149.18 & 298.85 & 0.02 \\
      26 & 116.36 & 86.58 & 245.99 & 566.03 & 0.00 \\
      27 & 128.58 & 98.63 & 263.32 & 587.30 & 0.00 \\
      28 & 149.23 & 144.81 & 279.49 & 493.05 & 0.00 \\ 
      29 & 145.72 & 93.24 & 231.98 & 335.98 & 0.00 \\
      30 & 167.50 & 144.27 & 378.89 & 874.99 & 0.00 \\
      31 & 169.30 & 152.83 & 404.40 & 997.82 & 0.00 \\
      32 & 174.32 & 152.89 & 362.63 & 570.41 & 0.00 \\
      33 & 212.59 & 237.60 & 410.05 & 923.25 & 0.00 \\      
      34 & 218.74 & 246.18 & 811.58 & 2243.79 & 0.00 \\
      35 & 221.93 & 230.44 & 613.64 & 1422.33 & 0.00 \\ 
      36 & 258.41 & 354.13 & 671.00 & 1333.50 & 0.00 \\
      37 & 258.95 & 342.97 & 947.95 & 2116.98 & 0.00 \\
      39 & 303.64 & 501.10 & 815.32 & 1373.29 & 0.00 \\
      40 & 293.24 & 649.42 & 884.32 & 1407.99 & 0.00 \\
      41 & 342.33 & 724.64 & 912.65 & 1517.10 & 0.00 \\
      43 & 274.39 & 267.95 & 1279.97 & 2194.84 & 0.00 \\
      45 & 277.26 & 414.19 & 1334.38 & 2470.13 & 0.00 \\
      47 & 311.91 & 405.1 & 1471.82 & 2172.13 & 0.00 \\      
      49 & 303.66 & 299.13 & 1487.65 & 2503.62 & 0.00 \\
      51 & 269.37 & 110.57 & 1571.46 & 2157.52 & 0.00 \\ 
      53 & 333.42 & 390.08 & 1804.47 & 2815.94 & 0.00 \\
      56 & 296.36 & 45.54 & 1895.23 & 3731.98 & 0.00 \\
      59 & 339.21 & 202.98 & 1765.18 & 3676.16 & 0.00 \\
      \hline
      Overall & & & & & 0.00 \\
      \hline
   \end{tabular}

   \caption{Number of messages needed to solve random target configurations targets in a field of 224 sensors using AWC and APO.}
   \label{tabtrackmess}
\end{center}
\end{table}

\section{Conclusions and Future Directions}
\label{summary}
In this article, we presented a new complete, distributed constraint
satisfaction protocol called Asynchronous Partial Overlay (APO).  Like
AWC, APO allows the agents to retain their autonomy because they can
obscure or completely hide internal variables and constraints.  In
addition, agents can refuse solutions posed by a mediator, instead
taking over as the mediator if for some reason they are unhappy with a
proposed solution.  We also presented an example of its execution on a
simple problem (section \ref{APO:example}) and proved the soundness
and completeness of the algorithm (section \ref{APO:proof}).  Through
extensive empirical testing on 10,250 graph instances from the graph
coloring and tracking domain, we also showed that APO significantly
outperforms the currently best known distributed constraint
satisfaction algorithm, AWC \cite{yokoo95awc}.  These tests have shown
that APO is better than AWC in terms of cycles to completion, message
usage, and runtime performance.  We have also shown that the runtime
characteristics can not be directly attributed to the speed of the
centralized solver.

APO's performance enhancements can be attributed to a number of
things.  First, APO exhibits a hill-climbing nature early in the
search which becomes more focused and controlled as time goes on.
Like other hill-climbing techniques this often leads to a satisfiable
solution early in the search.  Second, by using partial overlaying of
the information that the agents use in decision making, APO exploits
the work that has been previously done by other mediators.  This forms
a lock and key mechanism which promotes solution stability.  Lastly,
and most importantly, because APO uses dynamic, partial
centralization, the agents work on smaller, highly relevant portions
of the overall problem.  By identifying these areas of
decomposability, the search space can be greatly reduced which, in some
cases, improves the efficiency of the centralized search algorithm.

There are a vast number of improvements planned for APO in the future.
Probably the most important is to improve the centralized solver that
it uses.  In this article, an inefficient solver was chosen to show
the strengths of the distributed portions of APO.  We expect that
additional improvements in the algorithm's runtime performance can be
obtained by using a faster centralized search engine.  In addition,
modern solvers often use methods like graph reductions, unit
propagation and backbone guided search.  It is conceivable that
information gained from the centralized search engine could be used to
prune the domains from the variables for consistency reasons and
variables from the centralized subproblem for relevance reasons.  We
expect this will further focus the efforts of the agents additionally 
reducing the search time and communications usage of the algorithm.
 
Along with these improvements is the selective use of memory for
recording \emph{nogoods}.  Unlike AWC which uses the nogoods to ensure
a complete search, APO's completeness relies on one of the agents
centralizing the entire problem in the worst case.  Because of this
key difference, APO can be improved by simply remembering a small,
powerful subset of the nogoods that it discovers from mediation
session to session.  This would allow the algorithm to improve future
search by exploiting work that it had done previously.

What should be clear is that APO, and the cooperative mediation
methodology as a whole, opens up new areas for future exploration and
new questions to be answered in distributed problem solving.  We
believe that this work shows a great deal of promise for addressing a vast
number of problems and represents a bridge between centralized and
distributed problem solving techniques.

\acks{
Special thanks to Bryan Horling for his design and implementation of
the Farm simulation environment in which the experiment were run and
to Shlomo Zilberstein, Bart Selman, Neil Immerman, and Jose Vidal for
making numerous suggestions during the development of this work.
Lastly, the authors would like to thank the JAIR reviewers for their
helpful feedback and suggestions and Carlos Ansotegui and Jean-Charles
R\'egin for their lengthy discussion during the final revision to this
article.

The effort represented in this paper has been sponsored by the Defense
Advanced Research Projects Agency (DARPA) and Air Force Research
Laboratory, Air Force Materiel Command, USAF, under agreement number
F30602-99-2-0525.  The views and conclusions contained herein are
those of the authors and should not be interpreted as necessarily
representing the official policies or endorsements, either expressed
or implied, of the Defense Advanced Research Projects Agency (DARPA),
Air Force Research Laboratory, or the U.S. Government. The
U.S. Government is authorized to reproduce and distribute reprints for
Governmental purposes notwithstanding any copyright annotation
thereon.}


\end{document}